%% file: iclr2027_conference.tex
\documentclass{article}
\ifdefined\pdfminorversion\pdfminorversion=7\fi
\usepackage[preprint]{iclr2027_conference}
\usepackage{times}

\input{math_commands.tex}

\usepackage{hyperref}
\hypersetup{
    pdftitle={MS-TTRL: Breaking the Frequency Conflict in Multiple-Solution Test-Time Reinforcement Learning},
    pdfauthor={Zhixia Zhang, Binqian Xu, Huaiyang Wang, Yikun Ban, Dianbo Liu},
    pdfsubject={Preprint},
    hidelinks
}
\usepackage{url}
\usepackage{graphicx}
\usepackage{subcaption}
\usepackage{booktabs}
\usepackage{multirow}
\usepackage[table]{xcolor}
\usepackage{amsthm}
\usepackage{wrapfig}
\usepackage{listings}
\lstdefinestyle{prompt}{
    basicstyle=\ttfamily\small,
    breaklines=true,
    breakatwhitespace=true,
    columns=fullflexible,
    keepspaces=true,
    showstringspaces=false,
    frame=single,
    framesep=5pt,
    xleftmargin=6pt,
    xrightmargin=6pt,
    aboveskip=6pt,
    belowskip=8pt
}

\theoremstyle{definition}
\newtheorem{definition}{Definition}
\newcounter{msalgorithm}
\title{MS-TTRL: Breaking the Frequency Conflict in Multiple-Solution Test-Time Reinforcement Learning}

\author{
\textbf{Zhixia Zhang}\textsuperscript{1,}\thanks{Equal Contribution. Feel free to contact \href{mailto:zhixiaz@buaa.edu.cn}{\texttt{zhixiaz@buaa.edu.cn}} for details.}\quad
\textbf{Binqian Xu}\textsuperscript{2,}\footnotemark[1]\quad
\textbf{Huaiyang Wang}\textsuperscript{1} \\[0.3em]
\textbf{Yikun Ban}\textsuperscript{1}\quad
\textbf{Dianbo Liu}\textsuperscript{2} \\[0.6em]
\textsuperscript{1}Beihang University\quad
\textsuperscript{2}National University of Singapore \\[0.3em]
Code: \url{https://github.com/zzx-peter/Multiple-Solution-TTRL}
}

\begin{document}

\maketitle

\begin{abstract}
Scientific, engineering, and mathematical reasoning tasks can admit multiple valid solutions, while exhaustive annotation of the full solution set is often prohibitively costly. Test-time reinforcement learning (TTRL) applies reinforcement learning to unlabeled test instances, offering a label-free approach to multiple-solution reasoning. In this setting, generation frequency serves two opposing objectives: correctness estimation favors frequent outcomes, whereas outcome-level diversity preservation favors less frequent outcomes. We term this tension the \textbf{Frequency Conflict}. To resolve this conflict, we propose MS-TTRL, which obtains frequency-independent correctness estimates through Outcome-Level Self-Verification while reserving generation frequency solely for diversity preservation. Three-State Outcome Cache and the Candidate State support verification reuse across repeated observations and confidence-aware handling of ambiguous outcomes, while Confidence-Gated Inverse Probability Scaling selectively strengthens the learning signals of rare, reliably verified outcomes. Experiments across two benchmarks, four multiple-solution tasks, and three backbone models, each evaluated with five random seeds, show that MS-TTRL improves Recovery Rate by 19.13 percentage points on average over the best-performing label-free baseline and approaches the performance of the supervised IPS-GRPO.

\end{abstract}


\section{Introduction}




%

Reasoning over complex scientific, engineering, and mathematical problems is central to scientific discovery, mechanistic understanding, and the development of new solutions. Many of these problems are inherently underdetermined, with limited evidence or constraints admitting multiple valid explanations, structures, or solutions~\citep{chen2025hypospace}. Large language models are increasingly being explored as flexible solvers for such problems~\citep{li2025np, guo2025deepseek}. Reinforcement learning from verifiable rewards (RLVR), in particular, provides a direct way to adapt their behavior using verifiable task feedback~\citep{shao2024deepseekmath,hu2026open}. However, standard RLVR typically relies on external supervision, such as human annotations or oracle-provided correctness labels. In multiple-solution settings, the space of valid solutions can be large and heterogeneous, making exhaustive enumeration and annotation costly or even infeasible in practice. By deriving training signals from the model’s own generations, test-time reinforcement learning (TTRL) extends RLVR to unlabeled test instances~\citep{zuo2026ttrl}, making it an appealing paradigm for multiple-solution problems where complete external supervision is unavailable.

\begin{figure}[t]
    \centering
    \includegraphics[width=1.0\linewidth]{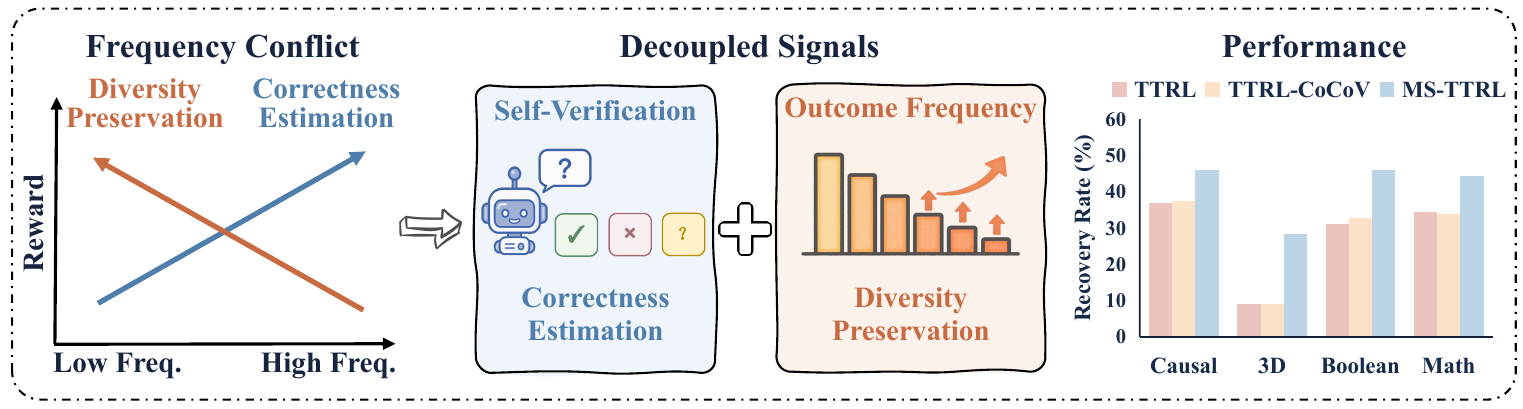}
    \caption{\textbf{Frequency conflict and signal decoupling in MS-TTRL.} \textbf{Left:} Frequency-based correctness estimation favors common outcomes, whereas diversity preservation favors rare ones. \textbf{Middle:} Self-verification estimates correctness, while outcome frequency promotes diversity. \textbf{Right:} MS-TTRL achieves higher Recovery Rate than TTRL and TTRL-CoCoV across four tasks.}
    \label{fig:idea_figure}
    \vspace{-12pt}
\end{figure}

However, existing TTRL methods have been developed primarily for single-solution reasoning, where training rewards are typically estimated from the frequency of model-generated answers~\citep{zuo2026ttrl,wang2026beyond,du2026distribution,yan2026if,mehrish2026core}. Specifically, multiple responses are sampled for each test instance, and the most frequent answer is treated as a pseudo-label to reward matching reasoning trajectories~\citep{zuo2026ttrl}. In single-solution tasks, this strategy aligns well with the task structure: different valid reasoning paths typically converge to the same correct answer~\citep{wang2022self}, allowing generation frequency to serve as a useful proxy for outcome correctness. Recent methods such as TTRL-CoCoV further promote diversity across reasoning trajectories while retaining frequency-based correctness estimation~\citep{li2026exploiting}. This remains compatible with single-solution reasoning because diverse trajectories can still converge to the same final outcome.

This compatibility breaks down in multiple-solution reasoning, where diversity must also be preserved across distinct valid outcomes~\citep{chen2025hypospace,sinha2026expected,puri2026reaching}. In this setting, generation frequency is forced to serve two opposing objectives: correctness estimation favors frequent outcomes, whereas outcome-level diversity preservation favors underrepresented ones~\citep{sinha2026expected,li2026beyond,anschel2025group}. Consequently, a low-frequency outcome receives conflicting interpretations: it may be regarded as unreliable by the correctness estimator, yet simultaneously prioritized for preserving outcome diversity. This ambiguity is fundamental because rare valid outcomes and rare erroneous outcomes may exhibit similarly low generation frequencies, making frequency alone insufficient to distinguish between them. We refer to this tension between correctness estimation and outcome diversity preservation as the \textbf{Frequency Conflict}.

The key to resolving the Frequency Conflict is to decouple correctness estimation from diversity preservation. We achieve this by using outcome-level self-verification as an independent correctness signal~\citep{weng2023large,hong2024closer,ma2025s2r}, while reserving generation frequency exclusively for measuring outcome rarity and diversity reweighting. Each distinct outcome is treated as a hypothesis to be verified: the model reasons backward from the candidate outcome, derives the conditions implied by it, and checks their consistency with the evidence, rules, and constraints of the problem.
The resulting verification pass rate, computed as the fraction of independent verification trials that accept an outcome, serves as a confidence estimate of its validity.
This design allows rare but reliably verified outcomes to receive stronger diversity-aware learning signals without similarly amplifying rare erroneous ones. Importantly, the novelty lies not in self-verification itself, but in using it to disentangle the two competing roles of generation frequency in multiple-solution TTRL. 
To the best of our knowledge, we are the first to systematically study the \textbf{Frequency Conflict} in label-free multiple-solution TTRL and to address it by explicitly decoupling correctness estimation from outcome diversity preservation.

Building on this decoupling principle, we propose Multiple-Solution Test-Time Reinforcement Learning (MS-TTRL), which integrates three components into a unified learning framework. MS-TTRL first obtains frequency-independent correctness estimates through \textit{Outcome-Level Self-Verification}. To make repeated verification efficient, \textit{Three-State Outcome Cache} stores and reuses verification results for previously observed outcomes. Verification uncertainty is handled by the \textit{Candidate State}, which retains ambiguous outcomes, assigns soft rewards according to verification confidence, and re-verifies them when the corresponding problem is revisited. Finally, \textit{Confidence-Gated Inverse Probability Scaling} strengthens the learning signals of rare but reliably verified outcomes while preventing uncertain or erroneous ones from being amplified.


Our main contributions are summarized as follows:

\textbf{[Problem Formulation].}
We identify the Frequency Conflict in multiple-solution TTRL, where generation frequency induces opposing preferences for correctness estimation and outcome diversity preservation, with frequent outcomes favored by correctness estimation and rare outcomes favored by diversity preservation.
To the best of our knowledge, we are the first to characterize this conflict and explicitly decouple these two objectives in this setting.


\textbf{[Algorithmic Innovation].}
We propose MS-TTRL, a Multiple-Solution TTRL framework that resolves the Frequency Conflict by using outcome-level self-verification for correctness estimation while reserving generation frequency exclusively for diversity preservation. To make this decoupling efficient and reliable in practice, MS-TTRL further supports verification reuse, confidence-aware treatment of ambiguous outcomes, and gated diversity reweighting.


\textbf{[Experimental Performance].}
We evaluate MS-TTRL on four multiple-solution tasks from two benchmarks, using three backbone models and five random seeds. Across all evaluated settings, MS-TTRL improves Recovery Rate by an average of \textbf{19.13} percentage points over the best label-free TTRL baseline, while achieving performance close to the supervised IPS-GRPO reference.
\section{Related Work}
Our work connects three lines of research. \textbf{Test-time reinforcement learning (TTRL).} TTRL optimizes policies on unlabeled test instances using self-generated supervision, typically constructing rewards from agreement among sampled responses. Existing methods primarily address single-solution reasoning, improving pseudo-label reliability and balancing exploration with exploitation~\citep{zuo2026ttrl,liu2025ettrl,li2026exploiting}.
\textbf{Trajectory-level diversity.} In single-solution reasoning, diversity-preserving methods use Pass@$K$ optimization and exploration-enhancing objectives to promote alternative reasoning trajectories leading to the same final answer~\citep{chen2025pass,peng2025simko,hu2026diversity}. Broader trajectory exploration, however, does not by itself ensure coverage of distinct valid outcomes.
\textbf{Outcome-level diversity.} In multiple-solution reasoning, outcome-level diversity methods promote distinct valid outcomes through inverse-probability scaling, distribution matching, or multi-answer objectives~\citep{sinha2026expected,li2026beyond,puri2026reaching}. These methods rely on reliable correctness signals to guide diversity optimization. We study multiple-solution TTRL without ground-truth labels or a reliable external verifier, decoupling correctness estimation from frequency-based diversity preservation to address the Frequency Conflict. Detailed related work appears in Appendix~\ref{sec:Detailed Related Work}.

\section{Frequency Conflict Between Correctness and Diversity}
\label{sec:Frequency Conflict Between Correctness and Diversity}

We consider test-time reinforcement learning for multiple-solution
problems. For each problem $x\in D$, let $\mathcal{Z}^{*}(x)$ denote
its finite set of valid outcomes. Given $G$ sampled responses
$\{y_i\}_{i=1}^{G}$, an outcome mapping $\phi$ induces the observed
outcome set $\mathcal{Z}(x)=\{\phi(y_i)\}_{i=1}^{G}$ and empirical
frequencies
\begin{equation}
    n(z\mid x)=\sum_{i=1}^{G}\mathbf{1}[\phi(y_i)=z],
    \qquad z\in\mathcal{Z}(x).
\end{equation}

\paragraph{Two roles of generation frequency.}
Frequency-based TTRL uses the most frequent outcome as a correctness
pseudo-label, whereas inverse-frequency weighting favors
underrepresented outcomes to preserve diversity. The corresponding
scores are
\begin{equation}
\begin{aligned}
    \text{Correctness estimation:}\quad
    r_{\mathrm{corr}}(z\mid x)
    &= \mathbf{1}[z=\hat z(x)],\\
    \text{Diversity preservation:}\quad
    r_{\mathrm{div}}(z\mid x)
    &= \frac{1}{n(z\mid x)},
\end{aligned}
\label{eq:frequency_roles}
\end{equation}
where
$\hat z(x)\in\arg\max_{z\in\mathcal{Z}(x)}n(z)$,
with ties broken by a fixed rule.
Here, $r_{\mathrm{corr}}$ scores pseudo-label agreement, while
$r_{\mathrm{div}}$ scores outcome rarity.
Inverse-frequency weighting promotes coverage when applied to
valid outcomes. Without validity labels, however, rarity alone
cannot distinguish a valid alternative from an error.

\begin{definition}[Frequency Conflict]
For a problem $x$, frequency-based correctness estimation and
diversity preservation exhibit a \emph{Frequency Conflict} if
there exist $z_a,z_b\in\mathcal{Z}(x)$ such that
\begin{equation}
\begin{gathered}
    n(z_a\mid x) > n(z_b\mid x),\\
    \underbrace{r_{\mathrm{corr}}(z_a\mid x) \ge
                r_{\mathrm{corr}}(z_b\mid x)}
               _{\text{correctness favors }z_a},
    \qquad
    \underbrace{r_{\mathrm{div}}(z_a\mid x) \le
                r_{\mathrm{div}}(z_b\mid x)}
               _{\text{diversity favors }z_b}.
\end{gathered}
\end{equation}
\end{definition}

For the scores in Eq.~\ref{eq:frequency_roles}, this conflict
arises whenever an observed outcome is less frequent than the
selected mode: it receives a zero correctness score but a higher
diversity score. Frequency alone cannot establish whether this
outcome is a valid but rare alternative that should be promoted or an error that should be suppressed. This ambiguity motivates estimating correctness independently of generation frequency. Section~\ref{sec:frequency_conflict_analysis} shows that extending pseudo-label selection to the top-$k$ outcomes can not alleviate this conflict.
\section{Method}
MS-TTRL resolves the Frequency Conflict by separating correctness estimation from frequency-based diversity optimization. It consists of three components: outcome-level self-verification, three-state outcome cache with an uncertainty buffer, and confidence-gated inverse probability scaling. Figure~\ref{fig:main_figure} illustrates how these components interact, while Algorithm~\ref{alg:ms_ttrl} in Appendix~\ref{sec:ms_ttrl_algorithm} provides the complete training procedure.
\begin{figure}
    \centering
    \includegraphics[width=1.0\linewidth]{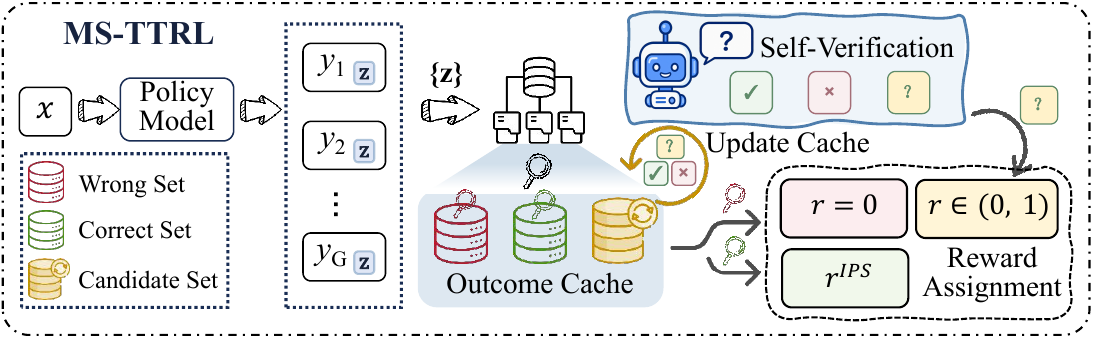}
    \caption{\textbf{Overview of MS-TTRL.} For each unlabeled problem, generated solution outcomes are self-verified and maintained in a three-state outcome cache. Their stored status is then combined with gated inverse-probability scaling to reward outcomes, thereby decoupling correctness estimation from diversity preservation with limited cost of self-verification.}
    \label{fig:main_figure}
    \vspace{-0.2cm}
\end{figure}

\subsection{Outcome-Level Self-Verification}

To resolve the Frequency Conflict, MS-TTRL estimates correctness through self-verification and reserves generation frequency for diversity preservation. Rare valid outcomes can thus receive positive correctness signals and diversity promotion.

The same model serves as actor and verifier, evaluating final outcomes rather than complete responses or intermediate reasoning. This shortens the context and avoids repeated verification of equivalent responses within a group.

To improve verification reliability, a reverse-reasoning prompt treats each outcome as a hypothesis: the verifier derives its implied conditions under the task rules and checks them against the problem. At step $t$, we map the $G$ responses sampled for $x$ to outcomes through $z=\phi(y)$ and merge duplicates to obtain

\begin{equation}
    \mathcal{Z}_t(x)=\{z_1,z_2,\ldots,z_M\},\qquad M\leq G.
\end{equation}

For each outcome $z$, we perform $n$ independent verification trials and define its verification pass rate as
\begin{equation}
    p_t(z\mid x)=\frac{1}{n}\sum_{i=1}^{n}
    \mathbf{1}\!\left[\operatorname{Verify}_i(z,x)=1\right],
\end{equation}
where $\operatorname{Verify}_i(z,x)=1$ denotes acceptance by trial $i$. The pass rate $p_t(z\mid x)$ serves as the correctness estimate throughout optimization.

\subsection{Three-State Outcome Cache}

Repeatedly verifying resolved outcomes is unnecessary and expensive. For each problem \(x\), MS-TTRL therefore maintains a persistent cache indexed by the problem--outcome pair \((x,z)\). The cache partitions all previously observed outcomes into a correct set \(\mathcal{S}_t(x)\), a candidate set \(\mathcal{C}_t(x)\), and a wrong set
\(\mathcal{W}_t(x)\). These sets represent verifier-assigned states rather
than ground-truth labels.

When problem \(x\) is revisited, only unresolved candidates and newly observed
outcomes require verification. The outcomes verified at step \(t\) are
\begin{equation}
    \mathcal{U}_t(x)
    =
    \mathcal{C}_{t-1}(x)
    \cup
    \left[
        \mathcal{Z}_t(x)
        \setminus
        \left(
            \mathcal{S}_{t-1}(x)
            \cup \mathcal{C}_{t-1}(x)
            \cup \mathcal{W}_{t-1}(x)
        \right)
    \right].
\label{eq:verification_query}
\end{equation}
Previously resolved correct and wrong outcomes retain their cached states without additional verifier calls.

For each \(z\in\mathcal{U}_t(x)\), its new state is determined by
\begin{equation}
    \operatorname{State}_t(z\mid x)
    =
    \begin{cases}
        \mathtt{correct},
        & p_t(z\mid x)\ge\tau_{\mathrm{high}},\\
        \mathtt{candidate},
        & \tau_{\mathrm{low}}
          < p_t(z\mid x) < \tau_{\mathrm{high}},\\
        \mathtt{wrong},
        & p_t(z\mid x)\le\tau_{\mathrm{low}},
    \end{cases}
\label{eq:state_assignment}
\end{equation}
where
\(0\leq\tau_{\mathrm{low}}<\tau_{\mathrm{high}}\leq1\).
The corresponding cache entry stores the assigned state and the latest
verification score. Correct and wrong outcomes are resolved and reused in
later steps, whereas candidates remain unresolved and are re-verified when
the problem is revisited.

\paragraph{Candidate state as an uncertainty buffer.}
A binary cache would have to accept or reject an outcome after a single verification round. Premature acceptance introduces erroneous positive rewards, whereas premature rejection may suppress rare valid outcomes before the verifier becomes sufficiently reliable. The candidate state postpones
this decision. Meanwhile, candidates retain policy support through their latest verification scores \(p_t(z\mid x)\): they are neither treated as confirmed correct outcomes nor assigned the zero reward used for wrong outcomes. The candidate state therefore acts as both a verification buffer and an exploration buffer.

\subsection{Confidence-Gated Inverse Probability Scaling}

Following IPS-GRPO, we scale the base reward to strengthen learning signals for infrequent correct outcomes and mitigate outcome-level collapse. Let \(n_t(z_i)\) be the count of outcome \(z_i\) in the current group, with empirical probability \(n_t(z_i\mid x)/G\). We restrict scaling to the correct set and control its strength with \(\lambda\) to prevent scaled rewards from overwhelming candidate soft rewards:
\begin{equation}
\tilde{r}_t(z_i)=
\begin{cases}
\lambda+(1-\lambda)\frac{G}{n_t(z_i\mid x)}, & z_i\in \mathcal{S}_t(x),\\
p_t(z_i\mid x), & z_i\in \mathcal{C}_t(x),\\
0, & z_i\in \mathcal{W}_t(x).
\end{cases}
\end{equation}

The coefficient \(\lambda\in[0,1]\) interpolates between full IPS (\(\lambda=0\)) and the unscaled base reward (\(\lambda=1\)).

This gating mechanism is particularly important in the label-free setting: applying inverse-probability scaling to candidate outcomes would amplify low-frequency verification noise. Candidate outcomes therefore retain policy support through soft rewards, whereas diversity amplification is restricted to outcomes assigned to the correct set.

\section{Experiments}
\subsection{Experimental Setup}
We evaluate MS-TTRL on three HypoSpace subsets~\citep{chen2025hypospace}---Causal Discovery, 3D Structure Reconstruction, and Boolean Expression Discovery---and on MACE-Math~\citep{wang2026evaluating}, an independent multiple-solution mathematical reasoning benchmark. Each problem admits a finite but potentially varying set of valid outcomes. Training uses only unlabeled problems; complete solution sets are reserved exclusively for evaluation.

All methods use the same evaluation budget, set to the largest valid-solution count in each task. Recovery Rate (RR) averages coverage across problems, whereas Outcome Recovery Rate (ORR) aggregates recovered outcomes over all valid outcomes and therefore gives more weight to problems with larger solution sets.

We compare against two label-free baselines: \textbf{TTRL}, which uses majority-vote pseudo-labels, and \textbf{TTRL-CoCoV}, which additionally promotes trajectory-level diversity. \textbf{IPS-GRPO} uses ground-truth correctness labels and serves only as a supervised reference.

HypoSpace experiments use Qwen3-1.7B, Qwen3-8B, and Phi4-4B-mini-Instruct; MACE-Math uses Qwen3-1.7B. 
We report means and standard deviations over five random seeds. Dataset splits and implementation details are provided in Appendices~\ref{sec:Dataset Details} and~\ref{sec:Training and Evolution Details}.
\subsection{Main Results}
\begin{table*}[t]
    \caption{
        Main results over 5 random seeds on the three HypoSpace subsets.
        RR and ORR denote Recovery Rate and Outcome Recovery Rate, respectively.
        Higher is better for both metrics.
        \textbf{IPS-GRPO, as shown in light gray, is a label-supervised reference rather than a baseline}.
        ``Best'' denotes the better-performing result between TTRL and TTRL-CoCoV for each metric.
    }
    \label{tab:main_results}
    \centering
    \footnotesize
    \setlength{\tabcolsep}{2.4pt}

    \begin{tabular}{@{}c*{6}{c}@{}}
        \toprule

        & \multicolumn{2}{c}{\textbf{Causal} (@16)}
        & \multicolumn{2}{c}{\textbf{3D} (@27)}
        & \multicolumn{2}{c}{\textbf{Boolean} (@20)} \\

        \cmidrule(lr){2-3}
        \cmidrule(lr){4-5}
        \cmidrule(lr){6-7}

        \smash{\raisebox{2.0ex}{\textbf{Method}}}
        & \makebox[1.15cm][c]{\textbf{RR(\%)} $\uparrow$}
        & \makebox[1.15cm][c]{\textbf{ORR(\%)} $\uparrow$}
        & \makebox[1.15cm][c]{\textbf{RR(\%)} $\uparrow$}
        & \makebox[1.15cm][c]{\textbf{ORR(\%)} $\uparrow$}
        & \makebox[1.15cm][c]{\textbf{RR(\%)} $\uparrow$}
        & \makebox[1.15cm][c]{\textbf{ORR(\%)} $\uparrow$} \\

        \midrule
        \multicolumn{7}{c}{\textbf{Qwen3-1.7B}} \\
        \midrule


        \textcolor{gray}{IPS-GRPO $^\dagger$}
        & \textcolor{gray}{$51.20 \pm 0.70$} & \textcolor{gray}{$28.19 \pm 0.53$}
        & \textcolor{gray}{$27.75 \pm 0.15$} & \textcolor{gray}{$16.35 \pm 0.04$}
        & \textcolor{gray}{$46.15 \pm 0.41$} & \textcolor{gray}{$37.38 \pm 0.57$} \\

        TTRL
        & $36.94 \pm 0.00$ & $17.42 \pm 0.00$
        & $9.07 \pm 0.00$  & $5.28 \pm 0.00$
        & $31.03 \pm 0.13$  & $22.63 \pm 0.13$ \\

        TTRL-CoCoV
        & $37.54 \pm 0.11$ & $17.63 \pm 0.04$
        & $9.07 \pm 0.00$  & $5.28 \pm 0.00$
        & $32.88 \pm 1.38$ & $25.11 \pm 0.74$ \\
        
        \textbf{MS-TTRL}
        & $45.89 \pm 0.49$ & $23.26 \pm 0.20$
        & $28.39 \pm 0.04$ & $16.93 \pm 0.03$
        & $46.00 \pm 1.10$ & $36.49 \pm 0.51$ \\

        $\Delta$(MS-TTRL$-$Best)
        & $\mathbf{+8.35}$ & $\mathbf{+5.63}$
        & $\mathbf{+19.32}$ & $\mathbf{+11.65}$
        & $\mathbf{+13.12}$ & $\mathbf{+11.38}$ \\
        \midrule
        \multicolumn{7}{c}{\textbf{Qwen3-8B}} \\
        \midrule
        
        
        \textcolor{gray}{IPS-GRPO $^\dagger$}
        & \textcolor{gray}{$65.23 \pm 0.69$} & \textcolor{gray}{$40.96 \pm 0.82$}
        & \textcolor{gray}{$54.51 \pm 0.58$} & \textcolor{gray}{$43.24 \pm 0.44$}
        & \textcolor{gray}{$46.46 \pm 1.41$} & \textcolor{gray}{$37.72 \pm 0.5$} \\
        
        TTRL
        & $36.94 \pm 0.00$ & $17.42 \pm 0.00$
        & $9.07 \pm 0.00$  & $5.28 \pm 0.00$
        & $26.03 \pm 0.80$ & $16.43 \pm 0.22$ \\
        
        TTRL-CoCoV
        & $36.91 \pm 0.00 $ & $17.42 \pm 0.00$
        & $9.48 \pm 0.27$  & $5.54 \pm 0.09$
        & $26.78 \pm 0.62$ & $19.28 \pm 0.45$ \\
        
        \textbf{MS-TTRL}
        & $61.65 \pm 0.44$ & $38.32 \pm 0.44$
        & $57.13 \pm 0.40$ & $45.35 \pm 0.63$
        & $47.95 \pm 0.53$ & $37.75 \pm 0.33$ \\

        $\Delta$(MS-TTRL$-$Best)
        & $\mathbf{+24.71}$ & $\mathbf{+20.90}$
        & $\mathbf{+47.65}$ & $\mathbf{+39.81}$
        & $\mathbf{+21.17}$ & $\mathbf{+18.47}$ \\
        \midrule
        \multicolumn{7}{c}{\textbf{Phi4-4B-mini-Instruct}} \\
        \midrule


        \textcolor{gray}{IPS-GRPO $^\dagger$}
        & \textcolor{gray}{$45.28 \pm 0.44$} & \textcolor{gray}{$23.44 \pm 0.26$}
        & \textcolor{gray}{$26.98 \pm 0.03$} & \textcolor{gray}{$15.73 \pm 0.02$}
        & \textcolor{gray}{$30.20 \pm 1.30$} & \textcolor{gray}{$21.08 \pm 1.04$} \\

        TTRL
        & $35.90 \pm 0.00$ & $16.29 \pm 0.00$
        & $0.64 \pm 0.00$  & $0.55 \pm 0.00$
        & $9.66 \pm 0.05$  & $6.69 \pm 0.13$ \\
        
        TTRL-CoCoV
        & $36.94 \pm 0.00$ & $17.42 \pm 0.00$
        & $1.36 \pm 0.13$ & $1.16 \pm 0.09$
        & $6.55 \pm 0.73$ & $5.69 \pm 1.06$ \\

        \textbf{MS-TTRL}
        & $42.03 \pm 0.10$ & $23.10 \pm 0.04$
        & $26.94 \pm 0.10$ & $15.55 \pm 0.16$
        & $26.10 \pm 1.26$ & $19.80 \pm 0.60$ \\

        $\Delta$(MS-TTRL$-$Best)
        & $\mathbf{+5.09}$ & $\mathbf{+5.68}$
        & $\mathbf{+25.58}$ & $\mathbf{+14.39}$
        & $\mathbf{+16.44}$ & $\mathbf{+13.11}$ \\
        \bottomrule
    \end{tabular}
\end{table*}
\begin{table}[tbp]
\centering
\footnotesize
\setlength{\tabcolsep}{3pt}
\caption{
Performance on the MACE-Math benchmark.
Higher is better for all metrics.
}
\label{tab:mace_math}
\begin{tabular}{lcc}
\toprule
\textbf{Method}
& \textbf{RR (\%) $\uparrow$}
& \textbf{ORR (\%) $\uparrow$} \\
\midrule


\textcolor{gray}{IPS-GRPO $^\dagger$}
& \textcolor{gray}{$68.19 \pm 0.49$}
& \textcolor{gray}{$55.26 \pm 0.57$} \\

TTRL
& $34.39 \pm 0.25$
& $22.75 \pm 0.21$ \\

TTRL-CoCoV
& $33.75 \pm 0.25$
& $22.34 \pm 0.12$ \\

\textbf{MS-TTRL}
& $44.26 \pm 0.56$
& $34.74 \pm 0.56$ \\

$\Delta$(MS-TTRL$-$Best)
& $\mathbf{+9.87}$
& $\mathbf{+11.99}$ \\

\bottomrule
\end{tabular}
\end{table}
MS-TTRL outperforms both label-free baselines in RR and ORR across all HypoSpace tasks and backbones (Table~\ref{tab:main_results}), with gains that remain consistent over five random seeds. On MACE-Math, it improves RR and ORR over the stronger label-free baseline by 9.87 and 11.99 percentage points, respectively (Table~\ref{tab:mace_math}), while remaining below the base model and supervised IPS-GRPO on both metrics.
Across the three HypoSpace tasks, MS-TTRL requires approximately 23.2\% more total training time than TTRL, with nearly unchanged peak GPU memory (Tables~\ref{tab:runtime} and~\ref{tab:gpu_peak}). It is faster than TTRL-CoCoV on Causal and 3D and matches its reported Boolean runtime.
High accuracy does not ensure broad outcome coverage: TTRL and TTRL-CoCoV approach perfect accuracy in several 3D settings while collapsing to a dominant outcome. TTRL-CoCoV does not consistently improve RR or ORR over TTRL, indicating that trajectory-level diversity alone does not prevent outcome-level collapse.


Accuracy results and further analysis appear in Appendices~\ref{subsec:Accuracy Report} and~\ref{sec:main_results_details}.





        

\begin{table}[h]
    \centering
    \caption{
    Training time and peak GPU memory on Qwen3-1.7B across three tasks.
    Lower is better.
    }
    \label{tab:efficiency}

    \begin{subtable}[t]{0.48\linewidth}
        \centering
        \caption{Runtime}
        \label{tab:runtime}
        \begin{tabular}{lccc}
            \toprule
            \textbf{Method}
            & \textbf{Causal}
            & \textbf{3D}
            & \textbf{Boolean} \\
            \midrule
            \textcolor{gray}{IPS-GRPO} & \textcolor{gray}{2h 25m} & \textcolor{gray}{39m} & \textcolor{gray}{32m} \\
            TTRL            & 2h 21m & 38m & 28m \\
            TTRL-CoCoV      & 3h 03m & 53m & 38m \\
            \textbf{MS-TTRL} & 2h 51m & 46m & 38m \\
            \bottomrule
        \end{tabular}
    \end{subtable}
    \hfill
    \begin{subtable}[t]{0.48\linewidth}
        \centering
        \caption{Peak GPU Memory (GB)}
        \label{tab:gpu_peak}
        \begin{tabular}{lccc}
            \toprule
            \textbf{Method}
            & \textbf{Causal}
            & \textbf{3D}
            & \textbf{Boolean} \\
            \midrule
            \textcolor{gray}{IPS-GRPO}
            & \textcolor{gray}{126.94}
            & \textcolor{gray}{121.44}
            & \textcolor{gray}{103.66} \\
            TTRL            & 129.14 & 117.57 & 102.87 \\
            TTRL-CoCoV      & 130.40 & 118.98 & 104.60 \\
            \textbf{MS-TTRL} & 131.83 & 123.29 & 107.54 \\
            \bottomrule
        \end{tabular}
    \end{subtable}
    \vspace{-0.3cm}
\end{table}
\subsection{Ablation and Sensitivity Analysis}
\subsubsection{Ablation Study}

We ablate three supporting components on Qwen3-1.7B: the cache across all three HypoSpace tasks, and the Candidate Set and Gated IPS on Causal. Appendix~\ref{sec:ablation_details} provides detailed interpretations.

\begin{figure}[h!]
    \centering
    \includegraphics[width=\linewidth]{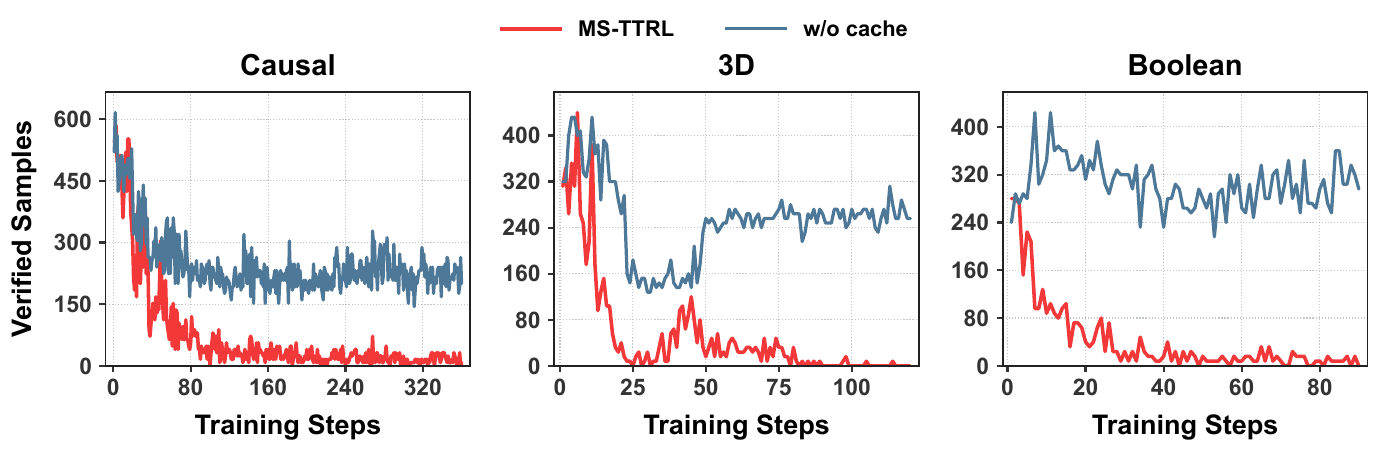}
    \caption{
    Ablation of the Three-State Outcome Cache on Qwen3-1.7B across the Causal, 3D, and Boolean tasks. Removing the cache mechanism substantially increases the number of verified samples throughout training, showing that reusing previously stored verification states effectively reduces repeated self-verification and lowers the additional inference cost.
    }
\label{fig:cache_sample_comparison}
\end{figure}

\textbf{Three-State Outcome Cache.}
Reusing resolved outcome states reduces self-verification samples by \textbf{75.9\%} overall across the three tasks and by \textbf{96.1\%} in the final epoch (Figure~\ref{fig:cache_sample_comparison}).

\textbf{Candidate-Based Uncertainty Buffer.}
Without the Candidate Set, outcomes receive binary labels using $\tau_{\mathrm{low}}$, $\tau_{\mathrm{high}}$, or their midpoint. On Causal, higher thresholds improve accuracy but reduce coverage, while lower thresholds produce the opposite trade-off (Table~\ref{tab:candidate_ablation}). Retaining uncertain outcomes allows MS-TTRL to achieve 93.75\% accuracy and 45.89\% RR, compared with 74.80\% and 41.79\% under the midpoint threshold.

\textbf{Confidence-Gated Inverse Probability Scaling.}
Replacing Gated IPS with ungated inverse-frequency scaling reduces Causal accuracy from 93.75\% to 59.59\% (Table~\ref{tab:candidate_ablation}). The gate protects correctness by restricting amplification to reliably verified outcomes; ungated scaling can amplify low-frequency errors that receive positive rewards.

\begin{table}[h!]
    \centering
    \setlength{\tabcolsep}{10pt}
\renewcommand{\arraystretch}{1.0}
    \caption{
    Ablation study of Gated IPS and the Candidate Set
    on Qwen3-1.7B for the Causal task.
    Results are averaged over five random seeds.
    }
    \label{tab:candidate_ablation}

    \begin{tabular}{cccc}
        \toprule
        \multirow{2}{*}{\textbf{Configuration}}
        & \multicolumn{3}{c}{\textbf{Causal}} \\
        \cmidrule(lr){2-4}
        & \textbf{ACC(\%)} $\uparrow$
        & \textbf{RR(\%)} $\uparrow$
        & \textbf{ORR(\%)} $\uparrow$ \\
        \midrule

        w/o Gated IPS
        & $59.59 \pm 0.68$
        & $49.46 \pm 0.79$
        & $27.68 \pm 0.54$ \\
    
        w/o Candidate Set (high)
        & $ 99.93 \pm 0.06 $
        & $ 37.15 \pm 0.28 $
        & $ 17.50 \pm 0.10 $ \\

        w/o Candidate Set (mid)
        & $74.80 \pm 0.90$
        & $41.79 \pm 0.77$
        & $20.55 \pm 0.49$ \\

        w/o Candidate Set (low)
        & $63.59 \pm 0.62$
        & $45.12 \pm 0.44$
        & $23.81 \pm 0.49$ \\
        
        \textbf{MS-TTRL}
        & $93.75 \pm 0.46$
        & $45.89 \pm 0.49$
        & $23.26 \pm 0.20$ \\

        \bottomrule
    \end{tabular}
\end{table}

\subsubsection{Hyperparameter Sensitivity Analysis}
We analyze sensitivity to key hyperparameters; Appendix~\ref{sec:hyperparameter_details} provides detailed interpretations.

\textbf{Number of self-verification trials.}
Increasing verification trials from 4 to 8 improves Causal recovery; 16 trials bring a smaller additional RR gain, although ORR improves further. On 3D, four trials already give recovery close to that with eight (Table~\ref{tab:verification_sensitivity}). We use 8 trials across tasks to balance verification reliability and cost.

\begin{table}[h]
    \centering
    \renewcommand{\arraystretch}{1.0}
    \caption{
    Sensitivity to the number of self-verification trials on
    Qwen3-1.7B for Causal and 3D Structure Reconstruction.
    }
    \label{tab:verification_sensitivity}
    \begin{tabular}{ccccc}
        \toprule
        \multirow{2}{*}{\textbf{Verification}}
        & \multicolumn{2}{c}{\textbf{Causal}}
        & \multicolumn{2}{c}{\textbf{3D}} \\
        \cmidrule(lr){2-3} \cmidrule(lr){4-5}
        & \textbf{RR (\%)} $\uparrow$
        & \textbf{ORR (\%)} $\uparrow$
        & \textbf{RR (\%)} $\uparrow$
        & \textbf{ORR (\%)} $\uparrow$ \\
        \midrule
        4  & 41.54 & 20.64 & 27.08 & 15.63 \\
        8  & 45.89 & 23.26 & 28.41 & 16.88 \\
        16 & 46.40 & 26.13 & 27.11 & 15.78 \\
        \bottomrule
    \end{tabular}
    \vspace{-0.3cm}
\end{table}

\par\medskip
\noindent
\begin{minipage}[t]{0.48\textwidth}
    \vspace{-0.1cm}
    \centering
    \small
    \setlength{\tabcolsep}{8pt}
    \captionsetup{type=table}
    \caption{
    Sensitivity to the thresholds for outcome-state assignment on
    Qwen3-1.7B for the 3D Structure Reconstruction task.
    One threshold is varied while the other is fixed.
    }
    \label{tab:threshold_sensitivity}
    \vspace{-0.1cm}
    \begin{tabular}{cccc}
        \toprule
        $\tau_{\mathrm{high}}$
        & $\tau_{\mathrm{low}}$
        & \textbf{RR (\%)} $\uparrow$
        & \textbf{ORR (\%)} $\uparrow$ \\
        \midrule

        \multicolumn{4}{c}{
        \textit{Varying $\tau_{\mathrm{low}}$
        ($\tau_{\mathrm{high}}=0.875$)}
        } \\
        \midrule

        0.875 & 0.375 & 27.18 & 15.83 \\
        0.875 & 0.500 & 28.41 & 16.88 \\
        0.875 & 0.625 & 21.32 & 12.29 \\

        \midrule
        \multicolumn{4}{c}{
        \textit{Varying $\tau_{\mathrm{high}}$
        ($\tau_{\mathrm{low}}=0.500$)}
        } \\
        \midrule

        0.750  & 0.500 & 27.16 & 15.83 \\
        0.875  & 0.500 & 28.41 & 16.88 \\
        1.000  & 0.500 & 26.14 & 15.67 \\

        \bottomrule
    \end{tabular}
\end{minipage}\hfill
\begin{minipage}[t]{0.495\textwidth}
\vspace{0pt}

\textbf{Outcome-state thresholds.}
Around the default threshold setting $(\tau_{\mathrm{high}}, \tau_{\mathrm{low}}) = (0.875, 0.5)$, recovery on the 3D task is less sensitive to changes in $\tau_{\mathrm{high}}$ than to increasing $\tau_{\mathrm{low}}$ to $0.625$ (Table~\ref{tab:threshold_sensitivity}). The latter leads to a substantial reduction in coverage, which is consistent with prematurely rejecting uncertain but potentially valid outcomes during outcome-state assignment.

\textbf{Strength of Gated IPS.}
Moderate scaling ($\lambda = 0.9$) achieves the best recovery performance on the 3D task among all tested values of the scaling parameter. Compared with this moderate setting, disabling IPS ($\lambda = 1$) reduces Recovery Rate from $28.39\%$ to $10.35\%$, while applying stronger scaling also results in lower recovery (Table~\ref{tab:lambda_sensitivity}).
\end{minipage}
\par\medskip

\begin{table}[h]
    \centering
    \vspace{-0.3cm}
    \caption{
    Sensitivity to the diversity scaling coefficient $\lambda$ on
    Qwen3-1.7B for the 3D Structure Reconstruction task.
    }
    \label{tab:lambda_sensitivity}
    \begin{tabular}{lccccccc}
        \toprule
        $\lambda$
        & 1.00
        & 0.95
        & 0.90
        & 0.85
        & 0.80
        & 0.75 \\
        \midrule
        \textbf{RR (\%)} $\uparrow$
        & 10.35 & 26.05 & 28.39 & 27.20 & 20.69 & 18.14 \\
        \textbf{ORR (\%)} $\uparrow$
        & 5.52 & 15.52 & 16.93 & 15.82 & 12.28 & 10.56 \\
        \bottomrule
    \end{tabular}
    \vspace{-0.4cm}
\end{table}

\subsection{Empirical Analysis of the Frequency Conflict}
\label{sec:frequency_conflict_analysis}
We test whether treating the top-$k$ most frequent outcomes as positive pseudo-labels resolves the Frequency Conflict. Table~\ref{tab:ttrl_topk} shows that increasing $k$ sharply degrades accuracy on Causal and 3D without reliably improving coverage, while Boolean exhibits a trade-off between lower accuracy and higher coverage.

Expanding the pseudo-label set admits both rare valid and rare invalid outcomes, so frequency-based pseudo-labels cannot reliably separate correctness from diversity. In contrast, the Self-Verify variant shows that simply decoupling correctness estimation from diversity preservation through \textit{Outcome-level Self-Verification} improves coverage over TTRL while maintaining high accuracy. These results support decoupling correctness estimation from frequency-based diversity weighting; detailed analysis appears in Appendix~\ref{sec:frequency_conflict_details}.
\begin{table*}[h]
    \centering
    \caption{
    Effect of extending TTRL from majority voting to the top-$k$ most frequent outcomes
    on Qwen3-1.7B across three multiple-solution tasks.
    TTRL-$k$ treats the top-$k$ generated outcomes as positive pseudo-labels.
    All methods use inverse-probability scaling when assigning rewards.
    }
    \label{tab:ttrl_topk}
    \footnotesize
    \setlength{\tabcolsep}{2.5pt}
    \begin{tabular}{lccccccccc}
        \toprule
        & \multicolumn{3}{c}{\textbf{Causal} (@16)}
        & \multicolumn{3}{c}{\textbf{3D} (@27)}
        & \multicolumn{3}{c}{\textbf{Boolean} (@20)} \\
        \cmidrule(lr){2-4}
        \cmidrule(lr){5-7}
        \cmidrule(lr){8-10}

        \textbf{Method}
        & \textbf{Acc(\%)} $\uparrow$
        & \textbf{RR(\%)} $\uparrow$
        & \textbf{ORR(\%)} $\uparrow$
        & \textbf{Acc(\%)}
        & \textbf{RR(\%)}
        & \textbf{ORR(\%)}
        & \textbf{Acc(\%)}
        & \textbf{RR(\%)}
        & \textbf{ORR(\%)} \\
        \midrule
        Self-Verify
        & 93.41 & 46.92 & 25.00
        & 97.57 & 27.10 & 15.78
        & 87.86 & 46.28 & 37.39 \\

        
        TTRL
        & 99.97 & 36.94 & 17.42
        & 99.87 & 9.07 & 5.28
        & 81.79 & 31.03 & 22.63 \\

        TTRL-2
        & 18.95 & 19.26 & 10.06
        & 22.10 & 2.81 & 2.19
        & 66.25 & 35.44 & 26.33 \\

        TTRL-3
        & 1.29 & 1.73 & 0.94
        & 0.00 & 0.00 & 0.00
        & 52.14 & 36.80 & 30.61 \\

        TTRL-4
        & 1.42 & 2.62 & 1.82
        & 2.68 & 1.71 & 1.52
        & 45.54 & 39.07 & 31.01 \\
        \bottomrule
    \end{tabular}
    \vspace{-0.3cm}
\end{table*}
\subsection{Learning Signals for Rare Valid Outcomes}
\label{sec:rare_valid_advantage}
On Qwen3-1.7B, we identify the least-frequent valid outcome(s) in each rollout group containing at least two distinct valid outcomes, retaining ties. Mean Advantage measures their average optimization signal, and Positive Advantage Rate measures the fraction receiving a positive signal.

\begin{figure*}[h]
    \centering
    \includegraphics[width=1\linewidth]{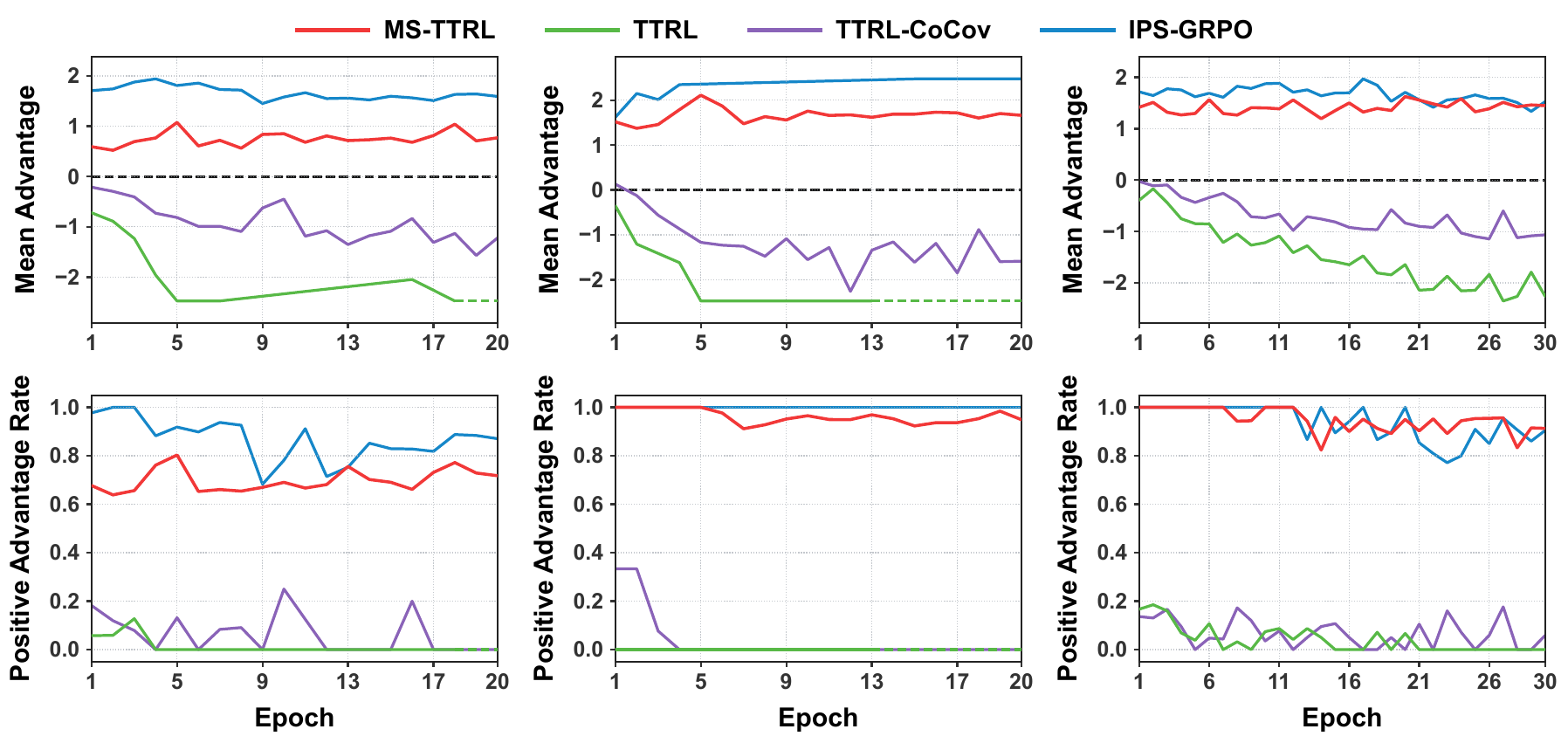}
    \caption{
    Learning signals assigned to rare valid outcomes on Qwen3-1.7B.
    For each rollout group containing at least two distinct valid outcomes, we identify the valid outcome(s) with the lowest generation frequency.
    Each subfigure reports the mean advantage and positive advantage rate across training epochs for one task.
    }
    \label{fig:rare_valid_advantage}
\end{figure*}

Across all three HypoSpace tasks, MS-TTRL maintains positive Mean Advantage and gives most rare valid outcomes positive advantages, whereas TTRL and TTRL-CoCoV tend to suppress them (Figure~\ref{fig:rare_valid_advantage}). Supervised IPS-GRPO also promotes these outcomes. This diagnostic is consistent with the proposed mechanism for preserving rare valid solutions; detailed definitions and discussion are provided in Appendix~\ref{sec:rare_valid_details}.

\section{Conclusion}
We presented MS-TTRL, a label-free test-time reinforcement learning framework for multiple-solution problems. We identified the Frequency Conflict caused by using generation frequency both to estimate correctness and to preserve outcome-level diversity. MS-TTRL separates these roles through outcome-level self-verification, Three-State Outcome Cache and Confidence-Gated Inverse-Probability Scaling. Across four tasks, three backbone models, and five random seeds, MS-TTRL consistently improves recovery over label-free baselines while approaching the supervised IPS-GRPO reference. These results show that decoupling correctness estimation from diversity preservation can improve valid-outcome coverage without requiring complete solution annotations during training.

\subsection*{AI use statement}

We used generative AI tools to polish and revise the language of the manuscript and to assist with code development. The core research ideas were developed by the authors without generative AI assistance. We manually reviewed the AI-assisted manuscript revisions and AI-generated code, and take full responsibility for the final content of this work, including its text and code.

\subsection*{Ethics statement}

This work studies test-time reinforcement learning on existing reasoning benchmarks. We do not identify any ethical concerns specific to the methods or experiments presented in this paper.

\subsection*{Reproducibility statement}

Our implementation is available at:
\begin{center}
\url{https://github.com/zzx-peter/Multiple-Solution-TTRL}
\end{center}
Appendices~\ref{sec:Dataset Details} and~\ref{sec:Training and Evolution Details} document the datasets, splits, training and evaluation configurations, and Appendix~\ref{sec:prompt_examples} provides the prompts used for Causal Discovery self-verification. The main experiments report means and standard deviations over five random seeds. Together, these materials support reproduction of the reported results.

\clearpage
\bibliography{iclr2027_conference}
\bibliographystyle{iclr2027_conference}
\newpage
\appendix
\section{MS-TTRL Algorithm}
\label{sec:ms_ttrl_algorithm}

Algorithm~\ref{alg:ms_ttrl} summarizes MS-TTRL. Outcome caches persist across visits to each problem, and only newly observed outcomes and cached candidates are self-verified.

\begin{center}
\begin{minipage}{\linewidth}
\refstepcounter{msalgorithm}\label{alg:ms_ttrl}
\hrule height 0.8pt
\vspace{4pt}
\textbf{Algorithm \themsalgorithm}\enspace Multiple-Solution Test-Time Reinforcement Learning
\vspace{4pt}
\hrule height 0.4pt
\vspace{5pt}
\textbf{Require:} Policy $\pi_\theta$; unlabeled problems $D$; rollout size $G$; verification trials $n$; thresholds $\tau_{\mathrm{high}},\tau_{\mathrm{low}}$; IPS coefficient $\lambda$; update budget $T$.
\par\vspace{4pt}
\begingroup
\small
\renewcommand{\arraystretch}{1.18}
\begin{tabular}{@{}r@{\hspace{0.6em}}l@{}}
 1: & Initialize $S(x)=C(x)=W(x)=\emptyset$ for every $x\in D$ \\
 2: & \textbf{for} $t=1,\ldots,T$ \textbf{do} \\
 3: & \quad Sample a batch $B\subseteq D$ \\
 4: & \quad \textbf{for each} $x\in B$ \textbf{do} \\
 5: & \qquad Sample $\{y_i\}_{i=1}^{G}\sim\pi_\theta(\cdot\mid x)$; extract $z_i=\phi(y_i)$ \\
 6: & \qquad $Z_t(x)\gets\{z_i\}_{i=1}^{G}$; $n_t(z)\gets\sum_{i=1}^{G}\mathbf{1}[z_i=z]$ \\
 7: & \qquad $U_t(x)\gets C(x)\cup\bigl(Z_t(x)\setminus(S(x)\cup W(x))\bigr)$ \\
 8: & \qquad \textbf{for each} $z\in U_t(x)$ \textbf{do} \\
 9: & \qquad\quad $p_t(z\mid x)\gets\frac{1}{n}\sum_{j=1}^{n}\mathbf{1}[\operatorname{Verify}_{\pi_\theta,j}(z,x)=1]$ \\
10: & \qquad\quad \textbf{if} $p_t(z\mid x)\ge \tau_{\mathrm{high}}$ \textbf{then} $S(x)\gets S(x)\cup\{z\}$; $C(x)\gets C(x)\setminus\{z\}$ \\
11: & \qquad\quad \textbf{else if} $p_t(z\mid x)\le\tau_{\mathrm{low}}$ \textbf{then} $W(x)\gets W(x)\cup\{z\}$; $C(x)\gets C(x)\setminus\{z\}$ \\
12: & \qquad\quad \textbf{else} $C(x)\gets C(x)\cup\{z\}$; store the latest $p_t(z\mid x)$ \\
13: & \qquad\quad \textbf{end if} \\
14: & \qquad \textbf{end for} \\
15: & \qquad \textbf{for} $i=1,\ldots,G$ \textbf{do} \\
16: & \qquad\quad \textbf{if} $z_i\in S(x)$ \textbf{then} $\tilde r_i\gets\lambda+(1-\lambda)G/n_t(z_i)$ \\
17: & \qquad\quad \textbf{else if} $z_i\in C(x)$ \textbf{then} $\tilde r_i\gets p_t(z_i\mid x)$ \\
18: & \qquad\quad \textbf{else} $\tilde r_i\gets0$ \\
19: & \qquad\quad \textbf{end if} \\
20: & \qquad \textbf{end for} \\
21: & \quad \textbf{end for} \\
22: & \quad Compute group-normalized advantages and update $\theta$ with GRPO on $B$ \\
23: & \textbf{end for} \\
24: & \textbf{return} $\pi_\theta$ \\
\end{tabular}
\endgroup
\par\vspace{5pt}
\hrule height 0.8pt
\end{minipage}
\end{center}

\section{Dataset Details}
\label{sec:Dataset Details}
\textbf{HypoSpace: Causal Discovery.}
The Causal Discovery task evaluates whether a model can recover multiple directed acyclic graphs (DAGs) that are consistent with the same intervention observations. Each problem contains four nodes, $\{A,B,C,D\}$, with up to six directed edges. Observations are generated by perturbing a subset of nodes and recording the resulting binary effects on the remaining nodes according to their causal descendants. Multiple DAGs may induce the same intervention observations and are therefore treated as distinct valid outcomes. Following the original HypoSpace construction, the complete set of admissible DAGs is exhaustively enumerated, enabling exact evaluation of solution coverage. We use 300 problems for training and 92 problems for evaluation. 

\textbf{HypoSpace: 3D Structure Reconstruction.}
The 3D Structure Reconstruction task requires the model to infer three-dimensional voxel structures from a two-dimensional top-view projection. Each problem is defined on a $3\times3$ grid with a maximum height of three layers. Multiple height assignments can correspond to the same top view, provided that they satisfy the physical support constraint that every voxel above the first layer is supported by a voxel directly below it. All valid height assignments consistent with the observed projection are treated as distinct outcomes. The finite solution space is exhaustively enumerated for each problem. We use 100 problems for training and 29 problems for evaluation.

\textbf{HypoSpace: Boolean Expression Discovery.}
The Boolean Expression Discovery task considers symbolic expressions over two variables, $x$ and $y$, using the operators \textsc{And}, \textsc{Or}, and \textsc{Not}. Candidate expressions are generated up to depth two and canonically deduplicated according to algebraic equivalences such as commutativity, idempotence, and associativity. Given a partial set of input--output observations, the model is asked to recover Boolean expressions consistent with the observed behavior. Since multiple expressions can satisfy the same observations, each problem naturally admits a finite set of valid outcomes. We use 50 problems for training and 28 problems for evaluation.

\textbf{MACE: Mathematical Reasoning.}
MACE is a multiple-answer benchmark designed to evaluate model behavior when a question admits several distinct correct answers. We use its mathematical reasoning subset, in which each problem asks the model to produce a number satisfying a specified mathematical property within a given interval, such as being a prime, square, cube, Fibonacci, or triangular number. The benchmark provides the complete set of valid answers for each problem, making it suitable for evaluating outcome-level coverage. For training, we select 250 problems with exactly six valid answers. For evaluation, we construct a separate test set by sampling 100 problems from each group corresponding to different numbers of valid answers, enabling us to assess the model's generalization across varying degrees of solution multiplicity.

To evaluate generalization and enable comparison with the supervised reference, all methods are trained on the same training split and evaluated on a held-out test split. During training, ground-truth labels are available only to IPS-GRPO and are withheld from all label-free methods.

\section{Training and Evaluation Details}
\label{sec:Training and Evolution Details}
All methods share the same training configuration unless otherwise specified.
For each prompt, we sample $G=8$ responses with temperature $1.0$ and top-$p=1.0$.
The maximum prompt and response lengths are both 1024 tokens.
Training uses AdamW with a learning rate of $1\times10^{-6}$.
The PPO clipping range is $[0.2, 0.28]$.
We normalize advantages by the group standard deviation following GRPO. The batch size is 16 prompts.
Models are trained for 20 epochs, except for Boolean, which uses 30 epochs. The default random seed is 1337.

For IPS-GRPO, rewards are inverse-frequency scaled using the empirical outcome frequency within each rollout group.
TTRL uses the most frequent outcome among the $G=8$ responses as the pseudo-label. For TTRL-CoCoV, we use Pass@$k$ with $k=4$, a diversity coefficient of $\lambda_{\mathrm{div}}=0.05$, confidence thresholds of $0.6/0.4$, and 8 verifier samples.

For MS-TTRL, we use Gated IPS with $\lambda=0.9$.
Outcome states are determined using verification thresholds
$\tau_{\mathrm{high}}=0.875$ and $\tau_{\mathrm{low}}=0.5$,
with 8 verification samples for each unresolved outcome.
Candidate outcomes receive their verification pass rate as a soft reward.

For evaluation, we sample $k$ responses for each problem with temperature $1.0$, where $k$ is set to the maximum number of valid solutions among all problems in the corresponding task. We report the average trajectory-level accuracy across sampled responses, together with solution coverage measured by Recovery Rate and Outcome Recovery Rate.
\section{Detailed Related Work}
\label{sec:Detailed Related Work}
\subsection{Test-Time Reinforcement Learning}

Test-time reinforcement learning (TTRL) adapts language models directly on unlabeled test instances, targeting settings in which ground-truth answers or reliable external reward signals are unavailable. Standard TTRL~\citep{zuo2026ttrl} samples multiple responses from the current policy, selects the majority outcome as a pseudo-label, and constructs rewards by comparing individual responses against this consensus. This enables policy improvement using only the model's own generations, but also makes the quality of optimization highly dependent on the reliability of the induced pseudo-supervision.

Subsequent work has primarily focused on improving the reliability and stability of this self-generated supervision. Several methods refine consensus-based pseudo-labeling through uncertainty estimation, confidence-aware filtering, selective supervision, distribution-aware aggregation, or interventions based on training dynamics~\citep{liu2025ettrl,wang2026beyond,du2026distribution,yan2026if,lin2026detecting,wu2025spine,xu2026hi}. Another line of work introduces self-verification or jointly improves generation and verification to identify unreliable consensus signals before they are used for optimization~\citep{pan2026coverrl,li2026exploiting,liao2026tool}. Despite these advances, generation frequency remains central to pseudo-label construction or confidence estimation in most existing TTRL methods. Moreover, these methods are predominantly developed for single-solution reasoning, where the main challenge is obtaining reliable supervision rather than maintaining coverage over multiple distinct valid outcomes. Our work instead studies TTRL in the multiple-solution setting, where correctness estimation must coexist with outcome-level diversity preservation.

\subsection{Trajectory-Level Diversity}

In conventional single-solution reasoning, different correct responses may follow substantially different reasoning trajectories while ultimately converging to the same correct outcome. Diversity in this setting therefore primarily concerns the coverage of alternative reasoning paths rather than distinct final answers. Prior studies have observed that reinforcement learning from verifiable rewards can improve Pass@1 while reducing Pass@K for larger $K$, suggesting that optimization may concentrate probability mass on a narrower subset of successful reasoning trajectories~\citep{chen2026does,yuan2026understanding}.

A growing body of work addresses this trajectory-level contraction by explicitly encouraging broader exploration during reinforcement learning. Representative approaches directly optimize Pass@K-style objectives, modify advantage estimation to account for multiple sampled trajectories, or introduce exploration-oriented objectives that discourage excessive concentration on a small number of high-reward reasoning paths~\citep{chen2025pass,walder2026pass,peng2025simko,hu2026diversity,he2026vane}. Related ideas have also been explored in the label-free setting. In particular, TTRL-CoCoV~\citep{li2026exploiting} augments TTRL with confidence-conditioned verification and exploration mechanisms to promote diversity among sampled reasoning trajectories. These methods improve exploration over alternative reasoning processes, but the resulting trajectories still converge to the same terminal outcome.

\subsection{Outcome-Level Diversity}

A different form of collapse arises when a problem admits multiple distinct valid outcomes. In this setting, reinforcement learning may concentrate probability mass on only a few easily discovered solutions, causing other valid outcomes to disappear from the policy distribution. The resulting objective is therefore not merely to maintain diverse reasoning trajectories, but to preserve coverage over the valid outcome space. Benchmarks such as HypoSpace provide controlled multiple-solution tasks in which such coverage can be explicitly evaluated through recovery-based metrics~\citep{chen2025hypospace, li2025np}.

Recent work has begun to address outcome-level collapse from several complementary perspectives. Frequency-aware methods such as IPS-GRPO and related approaches mitigate the reinforcement of dominant solutions by assigning stronger learning signals to underrepresented valid outcomes~\citep{sinha2026expected, anschel2025group}. Distribution-based methods such as DMPO align policy probabilities with reward-defined target distributions to promote broader coverage over high-reward outcomes~\citep{li2026beyond, gx2026kl}. Other approaches directly optimize diverse output generation, for example through set-level or multi-answer objectives~\citep{puri2026reaching,zhang20262,wang2026diversity}. Although these methods differ in optimization strategy, they share an important assumption: reliable correctness signals are available to determine which generated outcomes are valid before diversity is promoted.

Our setting removes this assumption. We study outcome-level diversity preservation when neither ground-truth solutions nor a reliable external verifier is available during training. In this setting, generation frequency simultaneously serves as a proxy for outcome correctness and as a signal of underrepresentation: high-frequency outcomes are favored by correctness estimation, whereas low-frequency outcomes are favored by diversity preservation. 
\section{Additional Experiment Results}
\label{sec:Additional Experiment Results}
\subsection{Accuracy Report}
\label{subsec:Accuracy Report}
As shown in Tables~\ref{tab:main_accuracy} and~\ref{tab:mace_math_accuracy}, we report the accuracy results corresponding to Tables~\ref{tab:main_results} and~\ref{tab:mace_math}. MS-TTRL maintains high accuracy while achieving strong valid-solution coverage.
\begin{table}[h!]
    \centering
    \caption{
    Accuracy over 5 evaluation seeds on the three HypoSpace subsets.
    Higher is better.
    }
    \label{tab:main_accuracy}
    \footnotesize
    \setlength{\tabcolsep}{6pt}
    \begin{tabular}{@{}lccc@{}}
        \toprule
        \textbf{Method}
        & \textbf{Causal ACC(\%)} $\uparrow$
        & \textbf{3D ACC(\%)} $\uparrow$
        & \textbf{Boolean ACC(\%)} $\uparrow$ \\
        \midrule

        \multicolumn{4}{c}{\textbf{Qwen3-1.7B}} \\
        \midrule


        \textcolor{gray}{IPS-GRPO $^\dagger$}
        & \textcolor{gray}{$95.14 \pm 0.63$}
        & \textcolor{gray}{$99.82 \pm 0.13$}
        & \textcolor{gray}{$88.54 \pm 0.79$} \\

        TTRL
        & $99.97 \pm 0.03$
        & $99.87 \pm 0.07$
        & $81.79 \pm 0.81$ \\

        TTRL-CoCoV
        & $97.73 \pm 0.18$
        & $99.97 \pm 0.05$
        & $77.79 \pm 0.91$ \\

        \textbf{MS-TTRL}
        & $93.75 \pm 0.46$
        & $99.13 \pm 0.53$
        & $90.54 \pm 0.99$ \\

        \midrule
        \multicolumn{4}{c}{\textbf{Qwen3-8B}} \\
        \midrule


        \textcolor{gray}{IPS-GRPO $^\dagger$}
        & \textcolor{gray}{$96.05 \pm 0.17$}
        & \textcolor{gray}{$98.62 \pm 0.34$}
        & \textcolor{gray}{$94.86 \pm 0.52$} \\

        TTRL
        & $99.92 \pm 0.08$
        & $99.87 \pm 0.11$
        & $84.50 \pm 0.79$ \\

        TTRL-CoCoV
        & $96.49 \pm 0.49$
        & $96.53 \pm 0.53$
        & $89.36 \pm 0.91$ \\

        \textbf{MS-TTRL}
        & $95.80 \pm 0.18$
        & $98.72 \pm 0.30$
        & $97.00 \pm 0.77$ \\

        \midrule
        \multicolumn{4}{c}{\textbf{Phi4-4B-mini-Instruct}} \\
        \midrule


        \textcolor{gray}{IPS-GRPO $^\dagger$}
        & \textcolor{gray}{$95.68 \pm 0.29$}
        & \textcolor{gray}{$99.77 \pm 0.15$}
        & \textcolor{gray}{$62.46 \pm 2.36$} \\

        TTRL
        & $99.92 \pm 0.10$
        & $10.34 \pm 0.00$
        & $38.00 \pm 0.13$ \\

        TTRL-CoCoV
        & $99.89 \pm 0.18$
        & $18.31 \pm 0.48$
        & $21.96 \pm 0.84$ \\

        \textbf{MS-TTRL}
        & $97.42 \pm 0.32$
        & $77.85 \pm 0.85$
        & $45.57 \pm 1.52$ \\

        \bottomrule
    \end{tabular}
\end{table}

\begin{table}[h]
    \centering
    \small
    \setlength{\tabcolsep}{7pt}
    \caption{
    Accuracy on the MACE-Math benchmark using Qwen3-1.7B.
    Higher is better.
    }
    \label{tab:mace_math_accuracy}
    \begin{tabular}{lc}
        \toprule
        \textbf{Method}
        & \textbf{ACC (\%) $\uparrow$} \\
        \midrule

        
        \textcolor{gray}{IPS-GRPO $^\dagger$}
        & \textcolor{gray}{$82.83 \pm 0.64$} \\
        
        TTRL
        & $65.99 \pm 0.29$ \\
        
        TTRL-CoCoV
        & $62.86 \pm 0.35$ \\

        \textbf{MS-TTRL}
        & $67.00 \pm 0.35$ \\

        \bottomrule
    \end{tabular}
\end{table}
\subsection{Training Dynamics}
\begin{figure}[t]
    \centering

    \begin{subfigure}[t]{0.33\textwidth}
        \centering
        \includegraphics[width=\linewidth]{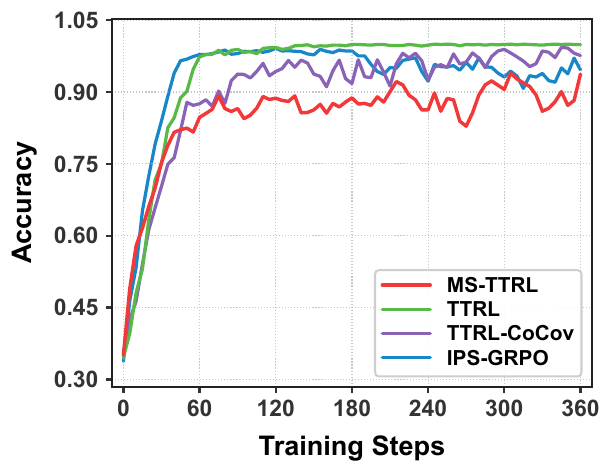}
    \end{subfigure}\hfill
    \begin{subfigure}[t]{0.33\textwidth}
        \centering
        \includegraphics[width=\linewidth]{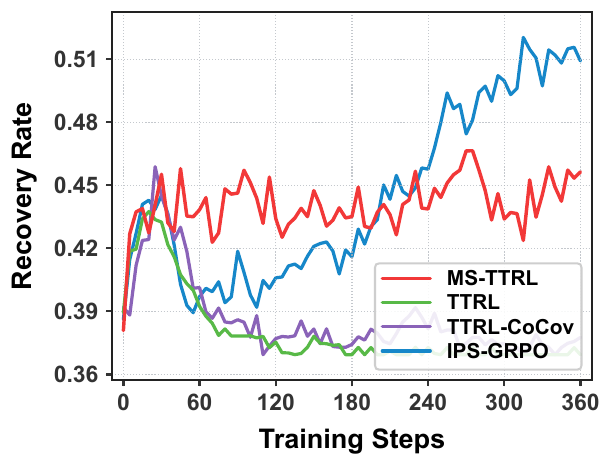}
    \end{subfigure}\hfill
    \begin{subfigure}[t]{0.33\textwidth}
        \centering
        \includegraphics[width=\linewidth]{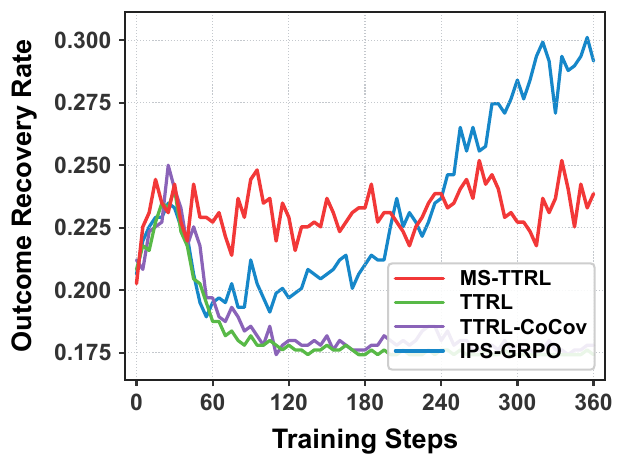}
    \end{subfigure}\hfill

    \vspace{0.5em}

    \begin{subfigure}[t]{0.33\textwidth}
        \centering
        \includegraphics[width=\linewidth]{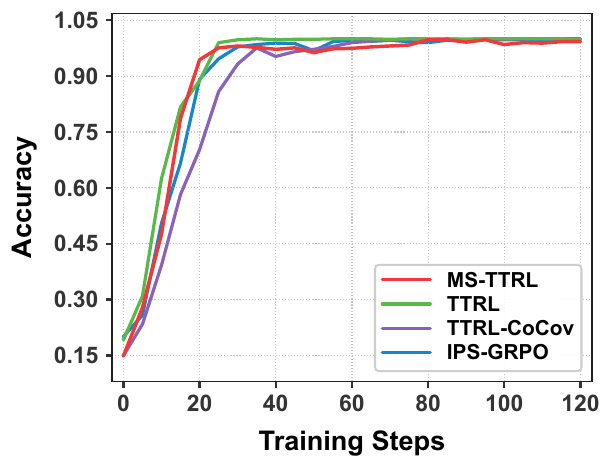}
    \end{subfigure}\hfill
    \begin{subfigure}[t]{0.33\textwidth}
        \centering
        \includegraphics[width=\linewidth]{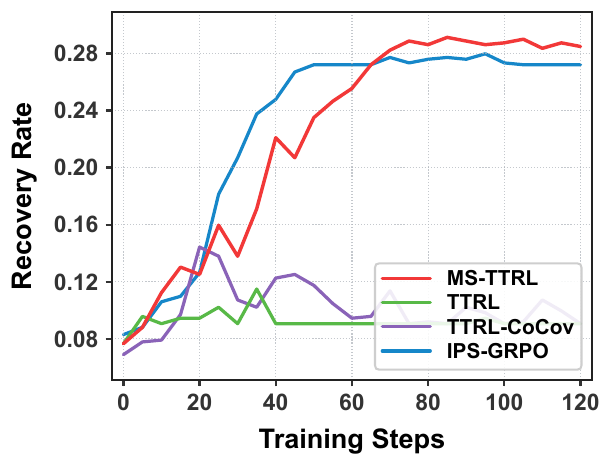}
    \end{subfigure}\hfill
    \begin{subfigure}[t]{0.33\textwidth}
        \centering
        \includegraphics[width=\linewidth]{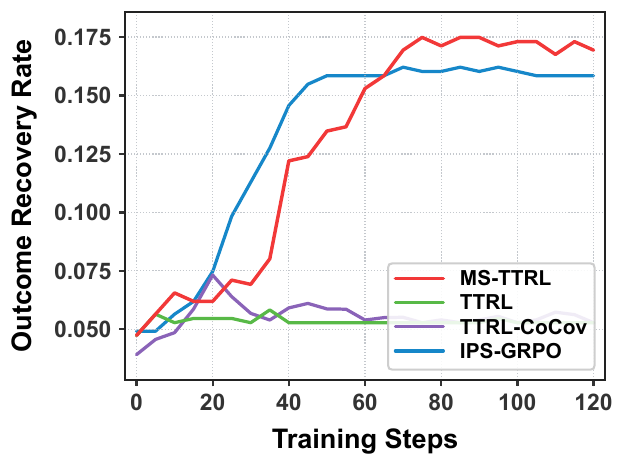}
    \end{subfigure}\hfill

    \begin{subfigure}[t]{0.33\textwidth}
        \centering
        \includegraphics[width=\linewidth]{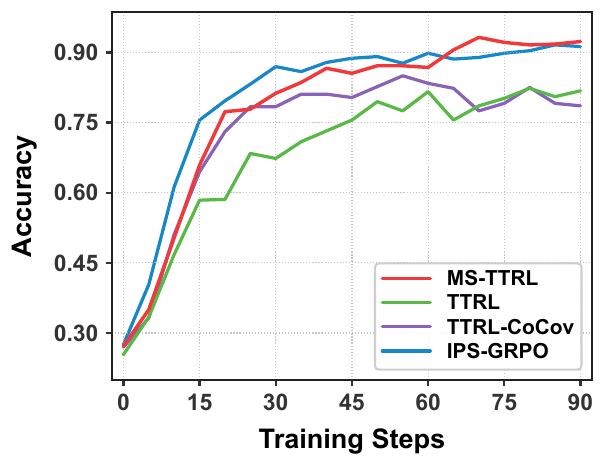}
        \caption{Accuracy}
        \label{fig:dynamics_acc}
    \end{subfigure}\hfill
    \begin{subfigure}[t]{0.33\textwidth}
        \centering
        \includegraphics[width=\linewidth]{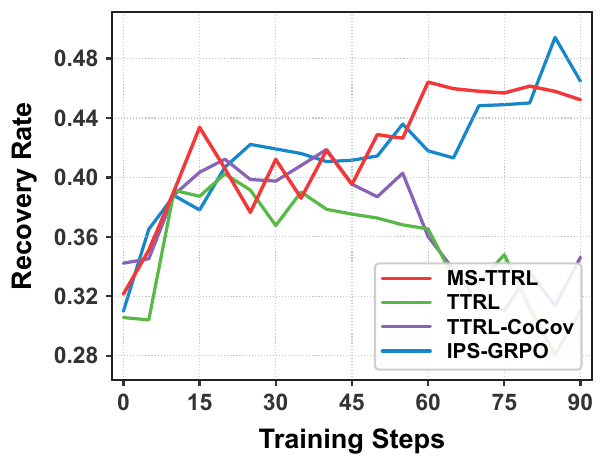}
        \caption{Recovery rate}
        \label{fig:dynamics_rr}
    \end{subfigure}\hfill
    \begin{subfigure}[t]{0.33\textwidth}
        \centering
        \includegraphics[width=\linewidth]{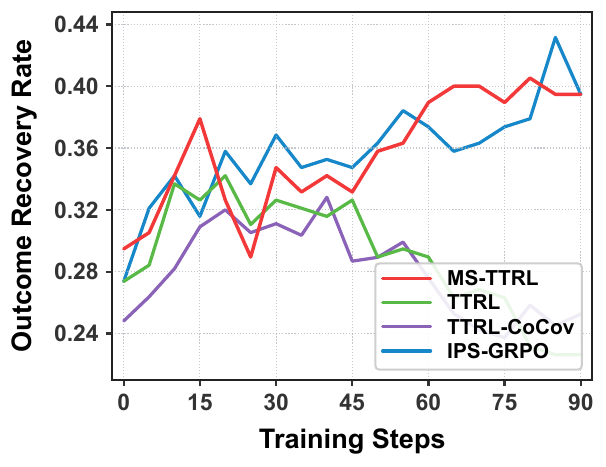}
        \caption{Outcome recovery rate}
        \label{fig:dynamics_mode_rr}
    \end{subfigure}\hfill
    
    \caption{Training dynamics on the Causal (top), 3D (middle), and boolean (bottom) tasks. From left to right: (a) accuracy, (b) recovery rate, (c) outcome recovery rate.}
    \label{fig:training_dynamics}
\end{figure}

We present the training curves of Qwen3-1.7B on three subtasks of the HypoSpace Benchmark: Causal, 3D, and Boolean. As shown in figure~\ref{fig:training_dynamics}, although TTRL and TTRL-CoCoV substantially improve accuracy, they also exhibit severe mode collapse. In particular, both the Recovery Rate (RR) and Outcome Recovery Rate (ORR) decline significantly during the later stages of training.
\section{Detailed Experimental Analysis}

\label{sec:detailed_experimental_analysis}

This appendix provides a detailed analysis supporting the experimental summaries in the main text. All associated tables and figures remain in the main text.

\subsection{Evaluation Metrics}

\label{sec:experiment_metrics_details}

For each subset, we determine the evaluation sampling budget based on the maximum number of valid solutions admitted by a single problem, and evaluate all methods using the same budget. We report the Recovery Rate and Outcome Recovery Rate to evaluate solution-space coverage. Recovery Rate first computes the fraction of valid solutions recovered for each problem and then averages across problems, assigning equal weight to every problem. Outcome Recovery Rate instead aggregates all recovered valid solutions across the dataset and divides by the total number of valid solutions, assigning equal weight to every solution and consequently more weight to problems with larger solution sets. Although the two metrics typically exhibit similar trends, they emphasize average per-problem coverage and aggregate coverage of the valid solution space, respectively.

\subsection{Main Results and Efficiency}

\label{sec:main_results_details}

As shown in Table~\ref{tab:main_results}, MS-TTRL consistently and substantially outperforms TTRL and TTRL-CoCoV across all three HypoSpace tasks in both Recovery Rate (RR) and Outcome Recovery Rate (ORR). On MACE-Math, MS-TTRL also achieves substantial improvements (Table~\ref{tab:mace_math}), indicating that the method generalizes beyond HypoSpace to an independent multiple-solution reasoning benchmark.

These gains persist across model scales (Qwen3-1.7B and Qwen3-8B), model families (Qwen and Phi), and five random seeds, suggesting that they are not tied to a particular backbone or favorable sampling randomness. Accuracy results are reported in Appendix~\ref{subsec:Accuracy Report}. Because our primary focus is valid-outcome coverage, the main table reports RR and ORR, while accuracy is used mainly for diagnostic analyses. Self-verification introduces additional inference cost, but the Three-State Outcome Cache reuses previously verified outcome states and keeps the overhead moderate. As shown in Table~\ref{tab:runtime}, MS-TTRL incurs approximately 23.2\% additional wall-clock training time over TTRL, remains faster than TTRL-CoCoV on Causal and 3D, and matches its reported runtime on Boolean. Its peak GPU memory allocation remains nearly unchanged relative to both methods (Table~\ref{tab:gpu_peak}), indicating that cache reuse amortizes repeated verification.

The results further reveal that high trajectory-level correctness does not necessarily imply broad solution-space coverage. TTRL and TTRL-CoCoV can achieve near-perfect trajectory-level accuracy in several settings while concentrating on only a small subset of valid outcomes. This collapse is particularly pronounced on the 3D Structure Reconstruction task: TTRL and TTRL-CoCoV with Qwen3-1.7B, and TTRL with Qwen3-8B, answer the problems correctly but effectively produce a single dominant outcome.

Improving trajectory-level diversity likewise does not necessarily improve outcome-level diversity. Although TTRL-CoCoV encourages exploration of alternative reasoning trajectories, its gains over vanilla TTRL in RR and ORR remain limited across most model-task configurations. Thus, diversifying reasoning paths toward the same answer does not by itself broaden coverage of distinct valid outcomes, motivating direct optimization at the outcome level.

MS-TTRL also remains competitive with supervised IPS-GRPO and occasionally achieves higher recovery in several settings, despite using no ground-truth solutions during training. One possible explanation is that HypoSpace tasks are relatively tractable for identifying valid solutions, while the main challenge is preserving diversity among them; under these conditions, exact correctness labels may provide limited additional benefit for recovery. MS-TTRL also assigns soft positive rewards to outcomes whose correctness is uncertain, some of which may be invalid. Although this introduces noise, it can preserve a broader exploratory distribution and occasionally improve solution-space coverage.

\subsection{Component Ablations}

\label{sec:ablation_details}

\paragraph{Three-State Outcome Cache.}

We first examine whether the Three-State Outcome Cache reduces the additional inference cost introduced by self-verification. As shown in Figure~\ref{fig:cache_sample_comparison}, reusing stored verification states substantially reduces repeated self-verification across all three tasks. The cache reduces the total number of self-verification samples by \textbf{75.9\%} over the three tasks, with the reduction reaching \textbf{96.1\%} in the final epoch. These results show that it amortizes self-verification by avoiding redundant checks of previously resolved outcomes.

\paragraph{Candidate-Based Uncertainty Buffer.}

We next ablate the Candidate-Based Uncertainty Buffer to study its role in preventing premature misclassification of uncertain outcomes. Without the Candidate Set, each outcome must be assigned an immediate binary correct/incorrect state using a single verification threshold. We consider three choices for this threshold: $\tau_{\mathrm{low}}$, $\tau_{\mathrm{high}}$, and their midpoint $(\tau_{\mathrm{high}}+\tau_{\mathrm{low}})/2$. As shown in Table~\ref{tab:candidate_ablation}, the midpoint causes a substantial accuracy drop on the Causal task. Raising the threshold improves accuracy but reduces RR and ORR, whereas lowering it has the opposite effect. A high threshold can falsely reject valid but insufficiently verified outcomes, causing them to lose positive rewards and reducing coverage. Conversely, a low threshold more readily accepts incorrect outcomes as valid, improving exploration at the cost of accuracy. The Candidate-Based Uncertainty Buffer avoids this forced trade-off by retaining ambiguous outcomes until additional verification evidence becomes available.

\paragraph{Confidence-Gated Inverse Probability Scaling.}

Finally, we evaluate the necessity of gating inverse-probability scaling by replacing Gated IPS with the original IPS rule, which applies inverse-frequency amplification without conditioning on the verification state. As shown in Table~\ref{tab:candidate_ablation}, removing the gate causes a sharp drop in Causal accuracy. In the label-free setting, a positive reward does not guarantee correctness: low-frequency incorrect outcomes can receive positive rewards and then be disproportionately amplified by vanilla IPS. This amplifies noisy rewards and can increase the policy probability of erroneous outcomes. Gated IPS prevents this failure mode by restricting amplification to outcomes verified with sufficiently high confidence, thereby promoting underrepresented valid outcomes without similarly amplifying uncertain or erroneous ones.

\subsection{Hyperparameter Sensitivity}

\label{sec:hyperparameter_details}

\paragraph{Number of self-verification trials.}

As shown in Table~\ref{tab:verification_sensitivity}, the verification budget needed for a reliable correctness estimate varies across tasks. On Causal Discovery, increasing the number of trials from 4 to 8 clearly improves RR and ORR; increasing it to 16 yields a smaller additional RR gain but further improves ORR. On 3D Structure Reconstruction, four trials already achieve near-optimal recovery. Thus, additional self-verification improves outcome-level correctness estimation, but its marginal benefit diminishes after a sufficient budget is reached. Because verification also incurs inference cost, we use 8 trials by default as a balance between reliability and efficiency.

\paragraph{Outcome-state thresholds.}

Table~\ref{tab:threshold_sensitivity} studies the two outcome-state thresholds, $\tau_{\mathrm{high}}$ and $\tau_{\mathrm{low}}$. Around the default setting of $\tau_{\mathrm{high}}=0.875$ and $\tau_{\mathrm{low}}=0.5$, varying $\tau_{\mathrm{high}}$ has only a minor effect on RR and ORR, indicating that assigning high-confidence outcomes to the Correct Set is relatively robust. Performance is more sensitive to $\tau_{\mathrm{low}}$: decreasing it to 0.375 causes little degradation, whereas increasing it to 0.625 noticeably reduces both recovery metrics. An overly large lower threshold may prematurely assign valid but temporarily uncertain outcomes to the Wrong Set. Because their cached states are then reused and their rewards become zero, such false rejections can progressively suppress their generation probability and reduce coverage. This asymmetry highlights the importance of retaining an intermediate Candidate Set.

\paragraph{Strength of Gated IPS.}

Finally, Table~\ref{tab:lambda_sensitivity} examines the Gated IPS coefficient $\lambda$. Setting $\lambda=1.0$ disables inverse-probability scaling and substantially reduces both RR and ORR, showing that correctness rewards alone are insufficient to prevent outcome-level collapse and that explicit amplification of underrepresented valid outcomes is important for recovery. Performance also deteriorates when $\lambda$ is too small, corresponding to overly aggressive scaling. Because IPS is applied only to Correct-Set outcomes, excessive scaling enlarges the reward gap between confirmed outcomes and uncertain Candidate outcomes, weakening the latter during exploration and potentially reducing diversity. Overall, $\lambda$ around 0.9 provides the best balance between reliable correctness signals and promotion of underrepresented valid outcomes.

\subsection{Frequency Conflict Analysis}

\label{sec:frequency_conflict_details}

A natural alternative to MS-TTRL is to retain frequency-based pseudo-labeling while expanding the positive set from the single majority outcome to the top-$k$ most frequent outcomes. We evaluate whether admitting more low-frequency outcomes resolves the Frequency Conflict between correctness estimation and outcome-level diversity preservation.

As shown in Table~\ref{tab:ttrl_topk}, enlarging the pseudo-label set does not reliably preserve both correctness and coverage; instead, it produces two failure modes.

First, Causal and 3D exhibit optimization collapse: accuracy is high at $k=1$ but drops rapidly as $k$ increases and approaches collapse for larger $k$. The expanded set treats more low-frequency invalid outcomes as positive, while inverse-probability scaling gives those outcomes stronger learning signals. Erroneous pseudo-labels are therefore both rewarded and amplified, allowing noisy supervision to dominate optimization.

Second, Boolean exhibits a clear accuracy--coverage trade-off. Small $k$ preserves higher accuracy but excludes more low-frequency valid outcomes; larger $k$ improves coverage by admitting more rare outcomes but also increases the chance of rewarding low-frequency invalid outcomes and thus degrades accuracy.

These results show that the Frequency Conflict cannot be resolved by simply increasing the number of pseudo-labels. As long as frequency determines both correctness and diversity rewards, changing $k$ only shifts the boundary between excluding rare valid outcomes and admitting rare invalid ones. Self-verification provides an independent correctness signal, while frequency can be reserved for diversity preservation. As a minimal realization, we perform $n$ independent verification trials per outcome and use the pass rate as its correctness reward; this decoupling already improves coverage while maintaining high accuracy.

\subsection{Learning Signals for Rare Valid Outcomes}

\label{sec:rare_valid_details}

We further investigate whether MS-TTRL promotes underrepresented valid outcomes in the current rollout distribution. We conduct this analysis on Qwen3-1.7B across the Causal, 3D, and Boolean tasks. For each training epoch, we consider problems whose rollout groups contain at least two distinct valid outcomes and identify, among the valid outcomes observed for each problem, the outcome with the lowest generation frequency.

We report Mean Advantage and Positive Advantage Rate. Mean Advantage is the average advantage assigned to underrepresented valid outcomes during policy optimization and indicates whether the update tends to promote or suppress them. Positive Advantage Rate is the fraction of rare valid outcomes receiving a positive advantage and measures how consistently they are promoted. We compare TTRL, TTRL-CoCoV, IPS-GRPO, and MS-TTRL; IPS-GRPO is included only as a supervised reference because it uses ground-truth validity labels.

As shown in Figure~\ref{fig:rare_valid_advantage}, MS-TTRL consistently provides positive learning signals to rare valid outcomes: its Mean Advantage remains positive, and most rare valid outcomes receive positive advantages. TTRL and TTRL-CoCoV, whose pseudo-labels are based on generation frequency, instead exhibit a clear suppressive effect. IPS-GRPO, which can directly identify valid outcomes using ground-truth labels, also consistently promotes underrepresented valid outcomes and therefore serves as a supervised reference upper bound.

These results show that self-verification identifies low-frequency valid outcomes without using generation frequency as a correctness proxy. This provides direct evidence that MS-TTRL maintains broader valid-outcome coverage than TTRL and TTRL-CoCoV under label-free training.

\section{Prompt Templates and Self-Verification Example}
\label{sec:prompt_examples}

We provide the prompts used in the Causal Discovery task to make the outcome-level self-verification procedure concrete. The first prompt specifies the task and asks the policy model to generate a candidate causal graph. The second prompt asks the verifier to check one candidate graph against the observations, and the final block shows a representative verifier response.

\subsection{Causal Task Input Prompt}
\begin{lstlisting}[style=prompt]
You are given observations from perturbation experiments on a causal system.

    Semantics:
    - When a node is perturbed, the perturbed node is 0.
    - A node is 1 if it is a downstream descendant of the perturbed node in the causal graph.
    - All other nodes are 0.

    Nodes: A, B, C, D
Constraint: The graph should have at most 6 edges.

    Observations:
    Perturb(B) -> A:0 B:0 C:0 D:0
    Perturb(C) -> A:0 B:1 C:0 D:1
    Perturb(D) -> A:0 B:1 C:0 D:0

    Task:
    Output a single directed acyclic graph (DAG) over the nodes above that explains all observations.
    Before producing the final answer, reason through the observations step by step.
    The final answer should follow the formatting rules.

    Formatting rules:
    1) Use only the listed nodes. No self-loops. No cycles.
    2) Respond with exactly one line:
    - If there are edges: Graph: A->B, B->C
    - If there are no edges: Graph: No edges
\end{lstlisting}

\clearpage
\subsection{Self-Verification Prompt}
\begin{lstlisting}[style=prompt]
Verify whether the candidate causal graph explains the perturbation observations.

Semantics:
- When a node is perturbed, the perturbed node is 0.
- A node is 1 if it is a downstream descendant of the perturbed node in the causal graph.
- All other nodes are 0.

Nodes: A, B, C, D

Constraint: The graph should have at most 6 edges.

Reasoning example:
An edge `X->Y` points from the upstream cause X to the downstream effect Y.
Follow arrows only in this left-to-right direction; do not reverse an edge.
For candidate graph `A->B, B->D`, B and D are descendants of A, while C is not.
Therefore, perturbing A gives A:0, B:1, C:0, D:1, so
`Perturb(A) -> A:0 B:1 C:0 D:1` is consistent. In contrast,
`Perturb(A) -> A:0 B:0 C:0 D:1` is inconsistent because B is reachable from A.

Perturbation observations:
Perturb(B) -> A:0 B:0 C:0 D:0
Perturb(C) -> A:0 B:1 C:0 D:1
Perturb(D) -> A:0 B:1 C:0 D:0

Candidate graph:
B->A, B->C, B->D, C->D

Reason through reachability for every listed perturbation. Judge only this candidate; multiple candidates may be valid.

First provide a reasoning paragraph. Then end your response with exactly one line:
Verification Result: True
or
Verification Result: False
\end{lstlisting}

\clearpage
\subsection{Representative Self-Verification Output}
\begin{lstlisting}[style=prompt]
To verify whether the candidate graph explains the perturbation observations, we need to check if the observed results match the expected values under the given causal graph.

The candidate graph is: **B->A, B->C, B->D, C->D**.

Let's analyze each perturbation:

1. **Perturb(B)**:
   - B is perturbed, so B:0.
   - B is a direct ancestor of A (B->A), so A is a downstream descendant of B. Thus, A:1.
   - C is a direct descendant of B (B->C), so C:1.
   - D is a descendant of C (C->D), so D:1.
   - Result: A:1, B:0, C:1, D:1.  
   - Observed result: A:0, B:0, C:0, D:0.  
   - **Inconsistent**. The observed result differs from what the graph predicts.

2. **Perturb(C)**:
   - C is perturbed, so C:0.
   - A is not a descendant of C (C->D, but A is not reachable from C), so A:0.
   - B is not a descendant of C, so B:0.
   - D is a descendant of C (C->D), so D:1.
   - Result: A:0, B:0, C:0, D:1.  
   - Observed result: A:0, B:1, C:0, D:1.
   - **Inconsistent**. The observed result differs from what the graph predicts.

3. **Perturb(D)**:
   - D is perturbed, so D:0.
   - A is not a descendant of D, so A:0.
   - B is not a descendant of D, so B:0.
   - C is not a descendant of D, so C:0.
   - Result: A:0, B:0, C:0, D:0.  
   - Observed result: A:0, B:1, C:0, D:0.  
   - **Inconsistent**. The observed result differs from what the graph predicts.

Since the perturbation of B results in inconsistent observations, the candidate graph does **not** explain the perturbation observations.

Verification Result: False
\end{lstlisting}

\end{document}

%% file: math_commands.tex
\usepackage{amsmath,amsfonts,bm}

\def\eqref#1{equation~\ref{#1}}

\def\1{\bm{1}}

\DeclareMathAlphabet{\mathsfit}{\encodingdefault}{\sfdefault}{m}{sl}
\SetMathAlphabet{\mathsfit}{bold}{\encodingdefault}{\sfdefault}{bx}{n}



%% file: iclr2027_conference.bib
@article{zuo2026ttrl,
  title={Ttrl: Test-time reinforcement learning},
  author={Zuo, Yuxin and Zhang, Kaiyan and Sheng, Li and Qu, Shang and Cui, Ganqu and Zhu, Xuekai and Li, Haozhan and Long, Xinwei and Hua, Ermo and Qi, Biqing and others},
  journal={Advances in Neural Information Processing Systems},
  volume={38},
  pages={131459--131483},
  year={2026}
}

@article{liu2025ettrl,
  title={Ettrl: Balancing exploration and exploitation in llm test-time reinforcement learning via entropy mechanism},
  author={Liu, Jia and He, ChangYi and Lin, YingQiao and Yang, MingMin and Shen, FeiYang and Liu, ShaoGuo},
  journal={arXiv preprint arXiv:2508.11356},
  year={2025}
}

@inproceedings{wang2026beyond,
  title={Beyond majority voting: Towards fine-grained and more reliable reward signal for test-time reinforcement learning},
  author={Wang, Weiqin and Wang, Yile and Chen, Kehao and Huang, Hui},
  booktitle={Proceedings of the 64th Annual Meeting of the Association for Computational Linguistics (Volume 1: Long Papers)},
  pages={37251--37265},
  year={2026}
}

@article{du2026distribution,
  title={Distribution-Aware Reward Estimation for Test-Time Reinforcement Learning},
  author={Du, Bodong and Huang, Xuanqi and Li, Xiaomeng},
  journal={arXiv preprint arXiv:2601.21804},
  year={2026}
}

@inproceedings{yan2026if,
  title={What if consensus lies? selective-complementary reinforcement learning at test time},
  author={Yan, Dong and Liang, Jian and Wang, Yanbo and Lu, Shuo and He, Ran and Tan, Tieniu},
  booktitle={Proceedings of the 64th Annual Meeting of the Association for Computational Linguistics (Volume 1: Long Papers)},
  pages={28957--28970},
  year={2026}
}

@article{lin2026detecting,
  title={Detecting and Mitigating the Correct-Answer Extinction Window in Test-Time Reinforcement Learning with Majority Voting},
  author={Lin, Hongxiang and Kuai, Zhirui and Xue, Erpeng and Wang, Lei},
  journal={arXiv preprint arXiv:2605.19444},
  year={2026}
}

@inproceedings{pan2026coverrl,
  title={Coverrl: Breaking the consensus trap in label-free reasoning via generator-verifier co-evolution},
  author={Pan, Teng and Yan, Yuchen and Wang, Zixuan and Zhang, Ruiqing and Hou, Guiyang and Zhang, Wenqi and Lu, Weiming and Xiao, Jun and Shen, Yongliang},
  booktitle={Proceedings of the 64th Annual Meeting of the Association for Computational Linguistics (Volume 1: Long Papers)},
  pages={29833--29853},
  year={2026}
}

@article{li2026exploiting,
  title={Exploiting Verification-Generation Gap: Test-Time Reinforcement Learning with Confidence-Conditioned Verification},
  author={Li, Jiahui and Shan, Jianfeng and Chen, Wenpei and Wu, Shunyu and Lou, Jian and Feng, Wenjie and Li, Dan and Ng, See-Kiong},
  journal={arXiv preprint arXiv:2606.03608},
  year={2026}
}

@article{chen2026does,
  title={Does reinforcement learning really incentivize reasoning capacity in llms beyond the base model?},
  author={Chen, Zhiqi and Lu, Rui and Zhao, Andrew and Wang, Zhaokai and Yue, Yang and Song, Shiji and Huang, Gao},
  journal={Advances in Neural Information Processing Systems},
  volume={38},
  pages={57654--57689},
  year={2026}
}

@article{yuan2026understanding,
  title={Understanding Diversity Collapse in RLVR via the Lens of Overtraining},
  author={Yuan, Suqin and Chen, Jinkun and Zheng, Jiyang and Li, Muyang and Feng, Lei and Wang, Dadong and Xiang, Tao and Liu, Tongliang and An, Bo},
  journal={arXiv preprint arXiv:2606.15455},
  year={2026}
}

@article{chen2025pass,
  title={Pass@ k training for adaptively balancing exploration and exploitation of large reasoning models},
  author={Chen, Zhipeng and Qin, Xiaobo and Wu, Youbin and Ling, Yue and Ye, Qinghao and Zhao, Wayne Xin and Shi, Guang},
  journal={arXiv preprint arXiv:2508.10751},
  year={2025}
}

@article{chen2025hypospace,
  title={HypoSpace: A Diagnostic Benchmark for Set-Valued Hypothesis Generation under Underdetermination and Sublinear Coverage Bounds},
  author={Chen, Tingting and Lin, Beibei and Yuan, Zifeng and Zou, Qiran and He, Hongyu and Goyal, Anirudh and Ong, Yew-Soon and Liu, Dianbo},
  journal={arXiv preprint arXiv:2510.15614},
  year={2025}
}

@article{sinha2026expected,
  title={Expected Return Causes Outcome-Level Mode Collapse in Reinforcement Learning and How to Fix It with Inverse Probability Scaling},
  author={Sinha, Abhijeet and Elango, Sundari and Liu, Dianbo},
  journal={arXiv preprint arXiv:2601.21669},
  year={2026}
}

@article{li2026beyond,
  title={Beyond Mode Collapse: Distribution Matching for Diverse Reasoning},
  author={Li, Xiaozhe and Li, Yang and Fang, Xinyu and Ding, Shengyuan and Li, Peiji and Chen, Yongkang and Ma, Yichuan and Lyu, Tianyi and Li, Linyang and Lin, Dahua and others},
  journal={arXiv preprint arXiv:2605.19461},
  year={2026}
}

@article{puri2026reaching,
  title={Reaching Beyond the Mode: RL for Distributional Reasoning in Language Models},
  author={Puri, Isha and Damani, Mehul and Shenfeld, Idan and Ghassemi, Marzyeh and Andreas, Jacob and Kim, Yoon},
  journal={arXiv preprint arXiv:2603.24844},
  year={2026}
}

@article{wu2025spine,
  title={Spine: Token-selective test-time reinforcement learning with entropy-band regularization},
  author={Wu, Jianghao and George, Yasmeen and Ye, Jin and Wu, Yicheng and Schmidt, Daniel F and Cai, Jianfei},
  journal={arXiv preprint arXiv:2511.17938},
  year={2025}
}

@article{xu2026hi,
  title={Hi-TTRL: Regulating Consensus with Hints for Test-Time Reinforcement Learning},
  author={Xu, Kunbin and Li, Xingzuo and Bai, Xuefeng and Chen, Kehai},
  journal={arXiv preprint arXiv:2608.03545},
  year={2026}
}

@article{liao2026tool,
  title={Tool verification for test-time reinforcement learning},
  author={Liao, Ruotong and R{\"o}hrich, Nikolai and Wang, Xiaohan and Zhang, Yuhui and Samadzadeh, Yasaman and Tresp, Volker and Yeung-Levy, Serena},
  journal={arXiv preprint arXiv:2603.02203},
  year={2026}
}

@article{walder2026pass,
  title={Pass@ k policy optimization: Solving harder reinforcement learning problems},
  author={Walder, Christian and Karkhanis, Deep Tejas},
  journal={Advances in Neural Information Processing Systems},
  volume={38},
  pages={152416--152445},
  year={2026}
}

@article{peng2025simko,
  title={Simko: Simple pass@ k policy optimization},
  author={Peng, Ruotian and Ren, Yi and Yu, Zhouliang and Liu, Weiyang and Wen, Yandong},
  year={2025}
}

@inproceedings{hu2026diversity,
  title={Diversity-incentivized exploration for versatile reasoning},
  author={Hu, Zican and Zhang, Shilin and Li, Yafu and Yan, Jianhao Elliott and Hu, Xuyang and Cui, Leyang and Qu, Xiaoye and Chen, Chunlin and Cheng, Yu and Wang, Zhi},
  booktitle={International Conference on Learning Representations},
  volume={2026},
  pages={71175--71205},
  year={2026}
}

@inproceedings{he2026vane,
  title={VANE: Guiding High-Value Exploration in RLVR via Outcome-Process Novelty Shaping},
  author={He, Xu and Guo, Jialiang and Xiong, Fucheng and Zhao, Haodong and Zeng, Ke and Cai, Xunliang and others},
  booktitle={Findings of the Association for Computational Linguistics: ACL 2026},
  pages={28721--28739},
  year={2026}
}

@article{li2025np,
  title={Np-engine: Empowering optimization reasoning in large language models with verifiable synthetic np problems},
  author={Li, Xiaozhe and Fang, Xinyu and Ding, Shengyuan and Li, Linyang and Duan, Haodong and Liu, Qingwen and Chen, Kai},
  journal={arXiv preprint arXiv:2510.16476},
  year={2025}
}

@inproceedings{anschel2025group,
  title={Group-aware reinforcement learning for output diversity in large language models},
  author={Anschel, Oron and Shoshan, Alon and Botach, Adam and Hakimi, Shunit Haviv and Gendler, Asaf and Baruch, Emanuel Ben and Bhonker, Nadav and Kviatkovsky, Igor and Aggarwal, Manoj and Medioni, Gerard},
  booktitle={Proceedings of the 2025 Conference on Empirical Methods in Natural Language Processing},
  pages={32382--32403},
  year={2025}
}

@inproceedings{gx2026kl,
  title={KL-regularized reinforcement learning for generative modelling is designed to mode collapse},
  author={GX-Chen, Anthony and Prakash, Jatin and Guo, Jeff and Fergus, Rob and Ranganath, Rajesh},
  booktitle={International Conference on Learning Representations},
  volume={2026},
  pages={50665--50693},
  year={2026}
}

@inproceedings{zhang20262,
  title={A $^2$ Search: Ambiguity-Aware Question Answering with Reinforcement Learning},
  author={Zhang, Fengji and Niu, Xinyao and Ying, Chengyang and Lin, Guancheng and Hao, Zhongkai and Fan, Zhou and Huang, Chengen and Keung, Jacky and Chen, Bei and Lin, Junyang},
  booktitle={International Conference on Learning Representations},
  volume={2026},
  pages={136394--136448},
  year={2026}
}

@inproceedings{wang2026diversity,
  title={Diversity-enhanced reasoning for subjective questions},
  author={Wang, Yumeng and Fan, Zhiyuan and Liu, Jiayu and Huang, Jen-tse and Fung, Yi R},
  booktitle={International Conference on Learning Representations},
  volume={2026},
  pages={75302--75332},
  year={2026}
}

@article{guo2025deepseek,
  title={DeepSeek-R1 incentivizes reasoning in LLMs through reinforcement learning},
  author={Guo, Daya and Yang, Dejian and Zhang, Haowei and Song, Junxiao and Wang, Peiyi and Zhu, Qihao and Xu, Runxin and Zhang, Ruoyu and Ma, Shirong and Bi, Xiao and others},
  journal={Nature},
  volume={645},
  number={8081},
  pages={633--638},
  year={2025},
  publisher={Nature Publishing Group UK London}
}

@article{shao2024deepseekmath,
  title={Deepseekmath: Pushing the limits of mathematical reasoning in open language models},
  author={Shao, Zhihong and Wang, Peiyi and Zhu, Qihao and Xu, Runxin and Song, Junxiao and Bi, Xiao and Zhang, Haowei and Zhang, Mingchuan and Li, YK and Wu, Yang and others},
  journal={arXiv preprint arXiv:2402.03300},
  year={2024}
}

@article{hu2026open,
  title={Open-reasoner-zero: An open source approach to scaling up reinforcement learning on the base model},
  author={Hu, Jingcheng and Zhang, Yinmin and Han, Qi and Jiang, Daxin and Zhang, Xiangyu and Shum, Heung-Yeung},
  journal={Advances in Neural Information Processing Systems},
  volume={38},
  pages={162239--162262},
  year={2026}
}

@article{mehrish2026core,
  title={CoRE: Consensus Rewards via Equilibrium for Test-Time Reinforcement Learning},
  author={Mehrish, Ambuj and Vascon, Sebastiano},
  journal={arXiv preprint arXiv:2608.09324},
  year={2026}
}

@article{wang2022self,
  title={Self-consistency improves chain of thought reasoning in language models},
  author={Wang, Xuezhi and Wei, Jason and Schuurmans, Dale and Le, Quoc and Chi, Ed and Narang, Sharan and Chowdhery, Aakanksha and Zhou, Denny},
  journal={arXiv preprint arXiv:2203.11171},
  year={2022}
}

@inproceedings{weng2023large,
  title={Large language models are better reasoners with self-verification},
  author={Weng, Yixuan and Zhu, Minjun and Xia, Fei and Li, Bin and He, Shizhu and Liu, Shengping and Sun, Bin and Liu, Kang and Zhao, Jun},
  booktitle={Findings of the Association for Computational Linguistics: EMNLP 2023},
  pages={2550--2575},
  year={2023}
}

@inproceedings{hong2024closer,
  title={A closer look at the self-verification abilities of large language models in logical reasoning},
  author={Hong, Ruixin and Zhang, Hongming and Pang, Xinyu and Yu, Dong and Zhang, Changshui},
  booktitle={Proceedings of the 2024 Conference of the North American Chapter of the Association for Computational Linguistics: Human Language Technologies (Volume 1: Long Papers)},
  pages={900--925},
  year={2024}
}

@inproceedings{ma2025s2r,
  title={S2r: Teaching llms to self-verify and self-correct via reinforcement learning},
  author={Ma, Ruotian and Wang, Peisong and Liu, Cheng and Liu, Xingyan and Chen, Jiaqi and Zhang, Bang and Zhou, Xin and Du, Nan and Li, Jia},
  booktitle={Proceedings of the 63rd Annual Meeting of the Association for Computational Linguistics (Volume 1: Long Papers)},
  pages={22632--22654},
  year={2025}
}

@article{wang2026evaluating,
  title={Evaluating and calibrating llm confidence on questions with multiple correct answers},
  author={Wang, Yuhan and Ni, Shiyu and Ding, Zhikai and Zhan, Zihang and Li, Yuanzi and Bi, Keping},
  journal={arXiv preprint arXiv:2602.07842},
  year={2026}
}
